\documentclass{article}

\PassOptionsToPackage{numbers}{natbib}

 \usepackage[main, final]{neurips_2026}

\usepackage[utf8]{inputenc} 
\usepackage[T1]{fontenc}    
\usepackage{hyperref}       
\usepackage{url}            
\usepackage{booktabs}       
\usepackage{amsfonts}       
\usepackage{nicefrac}       
\usepackage{microtype}      
\usepackage{xcolor}         

\usepackage{graphicx}
\usepackage{amsmath}
\usepackage{amsfonts}
\usepackage{multirow}
\usepackage{adjustbox}
\usepackage{caption}
\usepackage{subcaption}
\usepackage{wrapfig}
\usepackage{flafter}
\usepackage{makecell}        
\DeclareMathOperator*{\argmax}{arg\,max}

\usepackage{algorithm}
\usepackage{algorithmic}
\title{\textbf{MASTARS}: Multi-Agent Sequential Trajectory Augmentation with Return-Conditioned Subgoals}

\author{%
  \textbf{Jiwon Jeon}$^{1}$ \quad
  \textbf{Myungsik Cho}$^{1}$ \quad
  \textbf{Woojun Kim}$^{1}$ \quad
  \textbf{Seongmin Kim}$^{1}$ \\
  \textbf{Woohyeon Byeon}$^{1}$ \quad
  \textbf{Seungyul Han}$^{2*}$ \quad
  \textbf{Youngchul Sung}$^{1*}$ \\[0.4em]
  $^{1}$KAIST \quad $^{2}$UNIST \\[0.2em]
  {\small\texttt{\{jiwon.jeon, ms.cho, woojun.kim, seongmin.kim,}} \\
  {\small\texttt{woohyeon.byeon, ycsung\}@kaist.ac.kr}, \texttt{syhan@unist.ac.kr}}
}

\begin{document}

\maketitle

\begin{abstract}
The performance of offline Multi-Agent Reinforcement Learning (MARL) is limited by the quality of the fixed offline dataset, motivating trajectory augmentation. However, augmenting multi-agent trajectories presents a fundamental challenge of the diversity–coordination  trade-off: independent generation improves diversity but lacks coordination, while joint generation induces coordination but lacks diversity, reproducing the training distribution. To address this challenge in multi-agent data augmentation, we introduce MASTARS, a novel diffusion-based framework that generates multi-agent trajectories sequentially across agents while enforcing cross-agent consistent coordination via the method of inpainting. MASTARS produces trajectories that are diverse, coordinated and globally coherent. Experiments on various benchmarks such as MPE, SMAC, and SMAC-v2 demonstrate significant performance gains across multiple offline MARL methods. Code is available at \url{https://github.com/Jiwonjeon9603/MASTARS}.

\end{abstract}

\section{Introduction}\label{introduction}
\renewcommand{\thefootnote}{\fnsymbol{footnote}}\footnotetext[1]{Correspondence to: Seungyul Han (\texttt{syhan@unist.ac.kr}) or Youngchul Sung (\texttt{ycsung@kaist.ac.kr}).}\renewcommand{\thefootnote}{\arabic{footnote}}
Offline reinforcement learning (RL) \citep{Levine20offlineRLsurvey} aims to learn policies from pre-collected datasets without additional environment interaction, enabling safe and efficient policy development in domains where online exploration is costly or risky. 
While recent algorithmic advancements have yielded remarkable success \citep{Fujimoto21td3bc, Kostrikov22iql, Tarasov23rebrac, Mao23svr}, the effectiveness of these approaches remains dependent on the quality of the underlying dataset \citep{Schweighofer22dataset}.
To address this limitation, {\em data augmentation} has emerged as an important research direction in offline RL, and  several methods have been proposed based on the strong generative capabilities of diffusion models \citep{Ho20ddpm} to produce high-quality synthetic data in single-agent RL \citep{Janner22diffplan, Liang23adaptdiff, Lu23ser, He23mtdiff, Li24diffstitch, Lee24gta}.

Extending offline single-agent data augmentation to multi-agent RL (MARL) introduces unique challenges due to the combinatorial growth of joint state–action spaces and the need for agent coordination. While prior work has proposed specialized offline MARL methods \cite{Yang21icq, Pan22omar, Wang23omiga, Shao23cfcql, kim2025b3c}, their performance remains sensitive to the  quality and diversity of the dataset. 
Unlike single-agent augmentation, MARL requires modeling coordinated joint behaviors across agents. However, widely-considered independent per-agent diffusion models  neglect cross-agent dependencies, thereby hindering coordination. This  motivates  joint trajectory generation approaches \citep{yuan25madits, Oh24eaq, fuins}, but existing joint approaches   tend to produce limited diversity, frequently replicating the training distribution.

\textbf{The key challenge.}~Multi-agent augmentation is not merely a generative modeling problem; it is a constrained synthesis problem. A useful augmented trajectory must introduce {diverse} behavior, preserve inter-agent {coordination}, and remain globally {coherent} with the environment. We refer to this as the \emph{diversity--coordination--coherence trade-off}. Existing approaches sacrifice at least one of these requirements: independent generation improves diversity but breaks coordination, joint generation preserves coordination but often collapses toward the dataset distribution, and naive sequential generation can introduce incoherent global states.

To address these challenges  in multi-agent data augmentation, we propose \textbf{M}ulti-\textbf{A}gent \textbf{S}equential \textbf{T}rajectory \textbf{A}ugmentation with \textbf{R}eturn-conditioned \textbf{S}ubgoals \textbf{(MASTARS)}, a novel diffusion-based framework that generates cooperative multi-agent trajectories in an agent-wise sequential manner guided by high-value subgoals, while maintaining global coherence via a diffusion inpainting mechanism. 
MASTARS is built upon three core components: (i) an agent-wise sequential trajectory generation strategy, which generates coherent trajectories for each agent while coordinating with others via the key idea of \emph{inpainting}. 
To reduce the search space for trajectory augmentation, rather than synthesizing trajectories from scratch, MASTARS leverages value-based subgoals through:
(ii) partial generation based on subgoal identification, which preserves each agent's trajectory prefix up to its highest-value timestep and regenerates only the continuation, and (iii) subgoal-based generation ordering, which determines the order of sequential generation based on value-based subgoals.

Through these approaches, MASTARS  ensures diverse, coordinated, and coherent data augmentation in the multi-agent setting, enhancing  performance of major offline MARL methods when applied to the augmented dataset. Experiments on various MARL benchmarks, including the Multi-Agent Particle Environments (MPE) \citep{Lowe17mpe}, the StarCraft Multi-Agent Challenge (SMAC) \citep{Samvelyan19smac}, and SMAC-v2 \citep{ellis2023smacv2} demonstrate that datasets augmented by MASTARS significantly improve the performance of various offline MARL methods. 

The main contributions of this work are as follows:
\begin{itemize}
    \setlength{\leftskip}{-1.8em}
    \item We identify a diversity--coordination--coherence trade-off in multi-agent augmentation, where each of independent, joint and naive sequential strategies  sacrifices at least one of these properties. 
    \item We propose MASTARS, a plug-in diffusion-based framework resolving this tradeoff via inpainting-based sequential generation and reducing augmentation's search space with value-based subgoals.
    \item We demonstrate consistent performance gains of MASTARS across offline MARL algorithms on MPE, SMAC, and SMAC-v2, with ablations validating each component.
\end{itemize}
    
    

\section{Preliminaries}\label{preliminaries}

\subsection{Multi-Agent Offline Reinforcement Learning}\label{subsec:offline_marl}
We consider a fully cooperative Partially Observable Markov Game with $N$ agents, formalized as a Dec-POMDP $\langle N,\mathcal{S},\{\mathcal{A}^i\}_{i=1}^{N},\Omega,\mathcal{O},P,r,\rho,\gamma\rangle$ \citep{Oliehoek16decpomdp}. Here,  \(\mathcal{S} \) is the state space, \(\mathcal{A}^i \) the action space of agent $i$,  \(\Omega \) the observation space, $\rho:\mathcal{S}{\to}[0, 1]$ the initial state distribution, and $\gamma\!\in\![0,1)$ the discount factor.
At each timestep, agents share the environment state $s\!\in\!\mathcal{S}$ but receive private observations  $\{o^i\!\in\!\Omega\}$ according to the observation function $\mathcal{O}: \mathcal{S}{\times}\mathcal{N}{\to}\Omega $.
Each agent $i$ selects an action $a^i\!\in\!\mathcal{A}^i$, forming the joint action $\mathbf{a}\!=\!(a^1,{\ldots},a^N)$.  The environment transitions via $P(s'\mid s,\mathbf{a}): \mathcal{S}{\times}\mathcal{A}^1{\times}{\cdots}{\times}\mathcal{A}^N{\to}\mathbb{R}$ and yields a global reward  $r(s,\mathbf{a})$. Bold symbols denote joint quantities,
 e.g., $\boldsymbol{o} = (o^1,\ldots,o^N)$.

In the offline setting, we are given a fixed dataset  
$\mathcal{D}\!=\!\{\tau_m^{\text{full}}\}_{m=1}^{M}$ of $M$ trajectories  
$\tau^{\text{full}}=(\boldsymbol{o}_{0:T},\boldsymbol{a}_{0:T-1}, r_{0:T-1})$ with maximum trajectory length $T$, collected by an unknown behavior policy.  
No further interaction with the environment is permitted.  
The goal is to learn a decentralized set of policies
$\boldsymbol\pi=\{\pi^i(a^i\mid o^i;\theta^i)\}_{i=1}^{N}$ (or an equivalent centralized-training factorization) that maximizes the expected return
$
    J(\boldsymbol\pi)=\mathbb{E}_{\tau\sim\boldsymbol\pi,P}\!\Big[\sum_{t=0}^{T-1}\gamma^{t}r_{t}\Big]
$.

\subsection{Diffusion-Based Generative Models}\label{subsec:dm}
Diffusion models \citep{Ho20ddpm, Song21score} consist of a forward process that gradually corrupts data with Gaussian noise and a learned reverse process that denoises the corrupted data. In the forward process, noise is injected into $x_0 \sim q(x_0)$ over $K$ steps using a fixed variance schedule $\{\beta_k\}_{k=1}^K$: 
$
	q(x_k\mid x_{k-1}) = \mathcal{N}\bigl(\sqrt{1-\beta_k}x_{k-1},\beta_k I\bigr)
$,
resulting in a Gauss–Markov process with $q(x_k\mid x_0) = \mathcal{N}\bigl(\sqrt{\bar\alpha_k}x_0,(1-\bar\alpha_k)I\bigr)$, where $\bar\alpha_k = \prod_{i=1}^k(1-\beta_i)$. The posterior $q(x_{k-1}\mid x_k,x_0)$ is also Gaussian. The reverse generative model $p_\theta$ starts from $x_K{\sim}\mathcal{N}(0,I)$ and generates samples via the reverse transitions:
$
    p_\theta(x^{0:K}) = \mathcal{N}(x^K;0,I)\;\prod_{k=1}^K p_\theta(x^{k-1}\mid x^k).
$

DDPM \citep{Ho20ddpm} converts learning $p_\theta$ into learning a noise model $\epsilon_\theta$, enabling reverse sampling as $x_{k-1}{=}\frac{1}{\sqrt{\alpha_k}}\left(x_k - \frac{1 - \alpha_k}{\sqrt{1 - \bar{\alpha}_k}}\epsilon_\theta(x_k, k)\right){+}\sqrt{\beta_k}z$ with $x_K,z{\sim}\mathcal{N}(0,I)$. A neural network for the noise model $\epsilon_\theta(x_k,k)$ is trained to predict the noise $\epsilon \sim \mathcal{N}(0,I)$ in the forward decomposition $x^k = \sqrt{\bar\alpha_k}x_0 + \sqrt{1-\bar\alpha_k}\epsilon$, using the simple mean-squared error (MSE) objective:
\begin{align}
\hspace{-1em}
\label{eq:ddpm_loss}
    L_{\mathrm{diff}}^{\mathrm{DDPM}}(\theta)
= \mathbb{E}_{x_0\sim q(x_0),\,k\sim\mathcal{U}[1,K],\,\epsilon}
\bigl[\|\epsilon_\theta(x_k,k)-\epsilon\|^2\bigr].
\end{align}

\textbf{Diffusion Models in Data Augmentation.}~ To generate offline RL datasets, diffusion models \citep{Janner22diffplan, Liang23adaptdiff, Li24diffstitch, Lee24gta} learn the trajectory distribution $p_\theta(\tau)$ via $\epsilon_\theta$ from an offline dataset using Eq.~\eqref{eq:ddpm_loss}. The basic idea of RL dataset generation with diffusion models is described in Appendix~\ref{appendix:single_agent_diffusion_model}. In particular, Diffuser \citep{Ajay23cdiffuser} proposes a return-conditioned diffusion model with the loss \citep{Ho22cf-diffusion}:
\begin{equation}
\label{eq:rg_diff_loss}
\begin{aligned}
\mathcal{L}(\theta)
= \mathbb{E}_{k,\{\tau_0,R\}\in\mathcal{D},\epsilon,\beta{\sim}Bern(p)}\quad \Bigl[\bigl\|\epsilon
- \epsilon_\theta\bigl(\tau_k,k,(1-\beta)R + \beta\emptyset\bigr)
\bigr\|^2\Bigr],
\end{aligned}
\end{equation}
where $\emptyset$ is a null token for the unconditioned model. The reverse process uses the perturbed noise $(1+w)\epsilon_\theta(\tau_k,k,R) - w\epsilon_\theta(\tau_k, k, \emptyset)$, conditioned on return $R$ with guidance weight $w$. MADiff \citep{Zhu024madiff} generates sequences of observations by a return-conditioned diffusion model with an inter-agent attention architecture for MARL. We adopt MADiff as our base diffusion model and further train a reward function $g_\xi(\boldsymbol{o}, \boldsymbol{a})$, a transition model $f_\varphi(o^i, a^i)$, and an inverse dynamics model $I_\phi(o^i, {o^i}')$ to predict rewards, observations, and  actions from a generated observation sequence \citep{Ajay23cdiffuser, Li24diffstitch}. Our training objective for the auxiliary models is defined as
{ \small
\begin{equation}
\label{eq:dynamic_model_loss}
\begin{aligned}
    \mathcal{L}(\xi, \phi, \varphi) = \mathbb{E}_{\boldsymbol{o},\boldsymbol{a},r,\boldsymbol{o}' \sim \mathcal{D}} \Bigl[ \big(r - g_\xi(\boldsymbol{o},\boldsymbol{a})\big)^2 + \sum_{i=1}^{N} \Big(\big(a^i - I_\phi(o^i, {o^i}')\big)^2 + ({o^i}' - f_\varphi(o^i,a^i))^2\Big)\Bigr].
\end{aligned}
\end{equation}
}

\vspace{-0.5em}
\section{Related Works}\label{related_works}
\textbf{Offline MARL.}~
In offline MARL, the rapidly growing joint action space creates challenges such as coordination under partial observability, out-of-distribution (OOD) errors, and convergence to suboptimal policies. Prior works address these via conservative value estimation (CFCQL~\citep{Shao23cfcql}), hybrid optimization (OMAR~\citep{Pan22omar}), and global-to-local regularization (OMIGA~\citep{Wang23omiga}). 
While these methods focus on value-based constraints, diffusion models have shown efficacy in modeling complex action distributions and reducing extrapolation errors in single-agent offline RL.
Building on this, MADiff~\citep{Zhu024madiff} extends diffusion to multi-agent settings with attention-based coordination, and DoF~\citep{li2025dof} 
generalizes the Individual-Global-Max (IGM) principle to the Individual-Global-Distributed (IGD) framework,
aligning multi-agent outcomes with individual distributions.

\textbf{Sequential Modeling in MARL}.~
Recent work has explored treating MARL as a sequential generation problem \citep{mat, hu2025pmat, harl, seqcomm, dima, seqwm}. These approaches show that imposing an ordering over agents, communications, or latent variables can improve coordination and scalability. However, they  target policy learning and message passing rather than trajectory augmentation. In contrast, MASTARS adopts a sequential paradigm for data augmentation and combines it with diffusion-based inpainting to refine trajectories conditioned on other agents, enabling coordinated multi-agent synthesis.

\textbf{Data Augmentation.}~
Data augmentation has been widely explored in vision to improve generalization, with inpainting methods (e.g., RePaint \citep{lugmayr2022repaint}, PSM \citep{li2023psm}, CE \citep{pathak2016ce}) using generative models to reconstruct missing regions without retraining.
In offline RL, where performance is limited by the given dataset, augmentation strategies include model-based methods that learn uncertain dynamics \citep{kidambi2020morel,yu2020mopo,yu2021combo,zhang2023tatu} and generative approaches that synthesize high-quality trajectories \citep{Lu23ser,Li24diffstitch,Lee24gta}.

In multi-agent settings, EAQ \citep{Oh24eaq} and INS \citep{fuins} improve coordination through Q-guidance and sparse-attention-based synthesis. However, these joint generation methods often struggle with trajectory diversity.
While MADiTS \citep{yuan25madits} attempts to mitigate spatial imbalances by identifying underperforming agents and performing joint resampling via partial noising, its one-shot refinement process often lacks consistency under complex coordination, potentially leading to noisy target states and misaligned behaviors.
In contrast, MASTARS employs an agent-wise sequential framework with RePaint-style inpainting to ensure inter-agent harmonization, thereby synthesizing diverse, coherent, and coordinated trajectories that go beyond the limitations of the original data.

\begin{figure*}[t]
	\centering
    \vspace{-1.2em}
	\includegraphics[width=\linewidth,trim=37 3 55 17,clip]{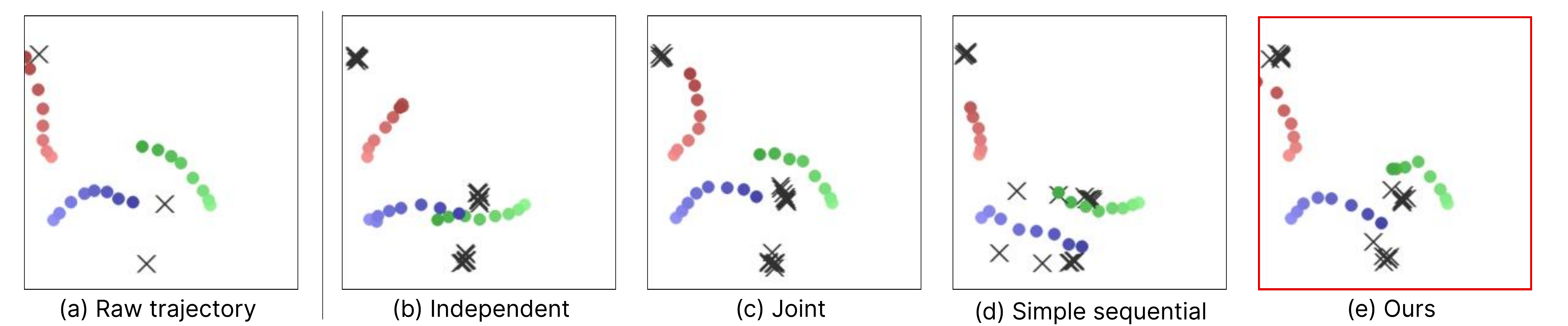}
    \vspace{-.2em}
	\caption{
 Comparison of trajectory generation strategies: Existing strategies struggle to jointly achieve coordination, diversity, and global coherence, while MASTARS satisfies all three. \texttt{X} marks are landmark positions reconstructed from each agent's observation.}
    	\label{fig:motivation}
\end{figure*}
\newpage
\section{Motivation}\label{sec:motivation}

Generating high-quality augmented trajectories for offline MARL requires accurate modeling of inter-agent dependencies and sufficient diversity to meaningfully expand the dataset’s support. However, existing diffusion-based data augmentation methods struggle to satisfy these requirements simultaneously. The limitation is  illustrated in Fig.~\ref{fig:motivation}, which compares 1) agent-wise independent generation, 2) joint generation, 3) naive sequential generation and 4) our method, all implemented using MADiff~\citep{Zhu024madiff}. Details of the generation processes are provided in Appendix~\ref{appendix:madiff_data_generation}.

We consider the offline Spread medium-replay task from the Multi-agent Particle Environment (MPE) \citep{Lowe17mpe}, where each of three agents (green, red, and blue) must reach a distinct landmark (marked as ‘\texttt{X}’) while avoiding collisions. Following the standard MPE setup, each agent's observation is a concatenation of its \texttt{own velocity}, its \texttt{own position}, the \texttt{landmark positions} encoded relative to the agent, and the \texttt{relative positions of the other agents}. A trajectory thus corresponds to a sequence of these per-agent observations, so a diffusion-based augmentation method must generate the entire sequence, including the landmark coordinates (\texttt{X}) at every timestep.  As shown in Fig.~\ref{fig:motivation}, none of the baseline strategies can consistently produce fully cooperative trajectories. The independent approach (Fig.~\ref{fig:motivation}(b)) frequently results in collisions because it neglects inter-agent dependencies. While the joint method mitigates collision risks, it lacks diversity by merely replicating the original data distribution, thereby inheriting the sub-optimality of the raw trajectories: Note that both blue and green agents are converging on the same landmark in Fig.~\ref{fig:motivation}(c).
Finally,  simple sequential generation (Fig.~\ref{fig:motivation}(d))  can yield diverse and collision-free trajectories through conditioning, but it fails to generate globally coherent trajectories; note that the generated rollouts exhibit inconsistent and fluctuating landmark positions although landmarks remain fixed throughout an episode. This inconsistency prevents agents from reliably reaching their designated goals.

Even this simple generation experiment reveals  the challenge of generating multi-agent trajectories that are simultaneously coherent, diverse, and well-coordinated across agents. 
In particular, ensuring global consistency over sequential generation is difficult to achieve via  naive agent-wise or sequential generation alone, motivating the need for complementary measures.

In the remainder of the paper, we propose a novel sequential generation framework that integrates subgoal-based partial generation with a diffusion-based {inpainting} mechanism to overcome the limitation of naive sequential generation. With these new ingredients, our approach enforces  consistency throughout sequential generation, and yields harmonized trajectories with coordination and global coherence. In our generation (Fig.~\ref{fig:motivation}(e)),  agents avoid collisions and  reach distinct landmarks while maintaining consistent landmark positions,  demonstrating that our method produces globally-coherent and well-coordinated trajectories that surpass the quality of the original offline dataset.

\section{Proposed Method}\label{method}
\begin{figure*}[t]
    \centering
    \vspace{-1.2em}
    \includegraphics[width=\linewidth]{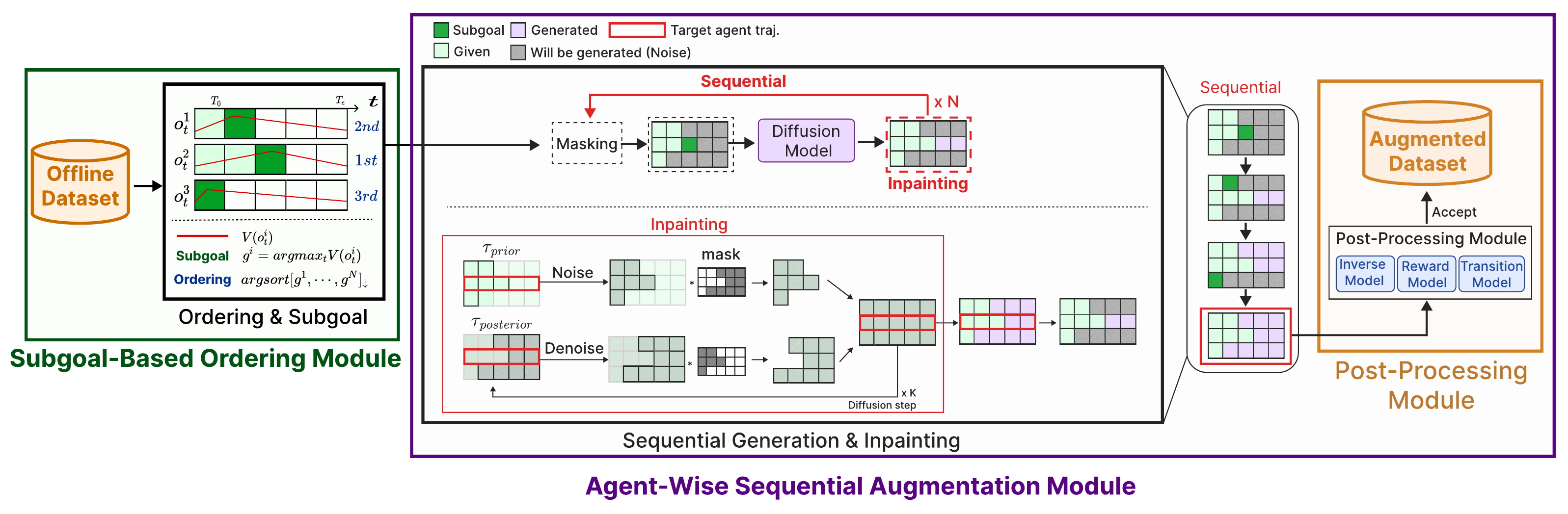}
    \vspace{-1.2em}
    \caption{
        \textbf{MASTARS overview.} Value-based subgoals are identified, agents are ordered by subgoal timing, and the selected segments are regenerated sequentially via diffusion inpainting with previously generated agents as known context.
        }    \label{fig:overflow_v1}
    \vspace{-1.2em}
\end{figure*}

To achieve desired multi-agent data augmentation, we combine two complementary mechanisms: i) resolving the diversity--coordination--coherence tradeoff via agent-wise sequential generation and ii) reducing the search space for trajectory augmentation via subgoal-based partial generation. Here, agent-wise sequential generation  promotes diversity and coordination by regenerating each agent conditioned on previously generated agents, while diffusion-based inpainting additionally enforces global coherence by treating previously generated agents and preserved trajectory segments as known regions. Subgoal-based partial generation preserves each agent's trajectory prefix up to its highest-value timestep and regenerates only the continuation, and subgoal-based ordering determines the generation order based on value-based subgoals. The overall flow of MASTARS is shown in Fig.~\ref{fig:overflow_v1}; the following subsections describe each component in detail.

\subsection{Agent-Wise Sequential Trajectory Generation}\label{method3:inpainting_module}

First, we formulate the agent-wise generation step as an inpainting problem~\citep{Elharrouss19inpaint}: previously generated agents and preserved segments form the known regions, and the current agent is generated as the missing region conditioned on them, ensuring coordination. To further enforce global coherence, we adopt RePaint~\citep{lugmayr2022repaint}, a training-free diffusion inpainting method whose iterative resampling and re-noising harmonize the generated agents with the global trajectory.

\paragraph{Inpainting-Based Agent-Wise Sequential Generation.} 

Formally, we define a MARL observation trajectory $\tau$ as an $N{\times}d{\times}H$ tensor, where $N$ denotes the number of agents, $d$ the observation dimension, and $H$ the time horizon.
Given an initial trajectory $\tau^0{=}\{\boldsymbol{o}_{t:t{+}H{-}1}\} \in \mathbb{R}^{N{\times}d{\times}H}$
and a generation order $(\kappa{(1)},{\ldots},\kappa{(N)})$,
we reconstruct each agent’s trajectory using the RePaint algorithm, treating previously generated agent trajectories as known regions. 
RePaint is applied sequentially to each agent, resulting in $N$ \emph{outer loop} invocations, 
each consisting of an \emph{inner loop} of $K$ denoising steps following the diffusion process in Section~\ref{subsec:dm}.
The initial trajectory $\tau^0$ is sampled directly from the dataset and is progressively refined through this RePaint-based reconstruction process.

At the $l$-th RePaint generation step, we start with $\tau^{l-1}$, the partially generated trajectory after the $l{-}1$ steps; that is, the trajectories for agents $\kappa(1), \ldots, \kappa(l{-}1)$ have already been generated. To identify the portion currently under generation, we define a binary mask $m^l$ of size $N{\times}d{\times}H$.
\begin{align}\label{eq:mask_in_Repaint}
\small
\hspace{-1em}
m^l(j,d,t) =
\begin{cases}
1~\forall t,d,  &\text{if } j \in \{\kappa(l),\ldots,\kappa(N)\},\\
0~\forall t,d, &\text{otherwise},
\end{cases}
\end{align}
where the segments of the remaining agents \(\kappa(l), \ldots, \kappa(N)\) after the \(l{-}1\) generation steps are set to 1, and the remaining segments to 0.

With the binary mask $m^l$, the reverse-sampling inner loop starts with index $k=K$.
For $k=K$, we  initialize
$\tilde \tau_{K}^{\text{posterior}}{\sim}\mathcal{N}\left(0,I\right)$ and
$\tilde \tau_{K}^{\text{prior}}{\sim}\mathcal{N}\left( \sqrt{\bar{\alpha}_K} {\tau}^{l{-}1}, (1{-}\bar{\alpha}_K) \mathbf{I} \right)$. Then, the denoising steps are applied according to the following sampling rules from $k{=}K$ to $k{=}1$ to obtain $\tilde{\tau}_0^l$:
\begingroup
\allowdisplaybreaks
\begin{subequations}\label{eq:repaint}
\begin{align}
\small
\begin{aligned}
\hspace{-1em}
\tilde \tau_{k}^{l}
= m^l \odot \tilde \tau_{k}^{\text{posterior}}
+ (1-m^l) \odot \tilde \tau_{k}^{\text{prior}} \in \mathbb{R}^{N \times d \times H}
\end{aligned}
\label{eq:repaint_c}
\\[-0.2em]
\begin{aligned}
\tilde\tau_{k-1}^{\text{posterior}} = \frac{1}{\sqrt{\alpha_k}}
\Bigl(
\tilde{\tau}_{k}^l
- \frac{1 - \alpha_k}{\sqrt{1 - \bar{\alpha}_k}}
\epsilon_\theta(\tilde{\tau}_{k}^l, k)
\Bigr) \quad + \sqrt{\beta_k}z,
\qquad z \sim \mathcal{N}(0,I)
\end{aligned}
\label{eq:repaint_b}
\\[-0.2em]
\begin{aligned}
\tilde \tau_{k-1}^{\text{prior}}
\sim \mathcal{N}
\Bigl(
\sqrt{\bar{\alpha}_{k}} {\tau}^{l-1},
(1 - \bar{\alpha}_{k})\, \mathbf{I}
\Bigr).
\end{aligned}
\label{eq:repaint_a}
\end{align}
\end{subequations}
\endgroup

Note that, for $k>1$, the prior input is a perturbed version of $\tau^{l-1}$ since $\bar{\alpha}_k < 1$, which helps improve coherence between the prior and posterior. Throughout  the reverse sampling loop for agents $\kappa(l),\cdots,\kappa(N)$, previously generated agents’ data are appended to enable coordinated generation, as shown in Eq.~\eqref{eq:repaint_c}.

After the internal denoising loop, the output of the $l$-th generation step is obtained as $\tau^{l} = \tilde \tau^{l}_0$, where the trajectory of the current agent $\kappa(l)$ is updated while those of agents $\kappa(1), \ldots, \kappa(l{-}1)$ remain unchanged. Although agents $\kappa(l{+}1), \ldots, \kappa(N)$ are temporarily processed to maintain multi-agent coherence, their outputs are discarded and regenerated in later steps. This inpainting procedure proceeds sequentially following the order ($\kappa(1), \ldots, \kappa(N)$) until all agent trajectories are generated.

\paragraph{Post-Processing and Validation for Generated Dataset.} 
After sequential generation, we obtain a synthesized observation trajectory $\tau = (\tilde{\boldsymbol{o}}_{t:t{+}H{-}1})$, where the tilde denotes generated quantities. 
To include this trajectory into an offline dataset, we infer the corresponding action and reward sequences using the pre-trained inverse dynamics model $I_\phi$ and reward model $g_\xi$, resulting in the following predictions:
$
    \tilde{a}_{t'}^i \!=\! I_\phi(\tilde{o}_{t'}^i, \tilde{o}_{t'+1}^i) 
$
and
$
    \tilde{r}_{t'} \!=\! g_\xi(\tilde{\boldsymbol{o}}_{t'}, \tilde{\boldsymbol{a}}_{t'}).
$

To ensure consistency with the environment dynamics and avoid out-of-distribution errors, we validate each generated trajectory using a pre-trained transition model $f_\varphi$. Trajectories are included in the augmented dataset only if their average prediction error satisfies the following condition:
\begin{align}\label{eq:sample_transition}
    \mathbb{E}_{i \in [1,N],\; t' \in [t,\, t+H-1]}
    \left\| \tilde{o}_{t'+1}^i - f_\varphi(\tilde{o}_{t'}^i, \tilde{a}_{t'}^i) \right\| < \epsilon,
\end{align}
where $\epsilon$ controls the acceptable level of deviation under the transition model \citep{Li24diffstitch}.

\subsection{Partial Generation via Subgoal Identification}\label{method3:subgoal}


While agent-wise sequential generation with inpainting addresses the diversity--coordination--coherence tradeoff, generating an entire trajectory from scratch  requires the diffusion model to search over the full trajectory space. To reduce this search space, we further use  \emph{subgoal-based partial generation}. That is, each agent's trajectory prefix up to a high-value timestep is preserved as an anchor, and only the remaining continuation is regenerated. This turns full-trajectory synthesis into a tractable continuation problem from this anchor.

Given an original trajectory $\tau^0 = \{\boldsymbol{o}_{t:t{+}H{-}1}\} \in \mathbb{R}^{N{\times}d{\times}H}$, we define a \emph{subgoal} $g^i$ for each agent $i$ as the timestep with the highest estimated value: $g^i = \argmax_{t' \in [t, t+H-1]} V_\zeta(o_{t'}^i).$

The prefix up to $g^i$ is treated as reliable behavior and kept fixed, while the suffix after $g^i$ is regenerated. This selective regeneration preserves useful behavior from the original dataset while reducing the search space for trajectory augmentation.

The subgoal-conditioned generation is integrated into our agent-wise sequential generation process by modifying the mask $m^l(j,d,t)$ in Eq.~\eqref{eq:mask_in_Repaint} as follows:
\begin{align}\label{eq:mask_in_Repaint_with_subgoal}
\small
\hspace{-1em}
m^l(j,d,t) =
\begin{cases}
1~\forall d,  &\text{if } j \in \{\kappa(l),\ldots,\kappa(N)\},~ t > g^j, \\
0~\forall t,d, &\text{otherwise}.
\end{cases}
\end{align}

\subsection{Agent Ordering to Preserve Coordination Flexibility}\label{method2:agnet_ordering}
\begin{table*}[t]
	\small
    \vspace{-1.2em}
    \caption{Normalized return in the MPE environment (\textbf{bold}: best, \underline{underline}: second best)}
    \vspace{-0.4em}
    \label{table:mpe_main_result}
	\centering
    \renewcommand{\arraystretch}{1.15}
    \robustify\bfseries
    \resizebox{\textwidth}{!}{%
      \setlength{\tabcolsep}{4pt}\begin{tabular}{@{}l||l|ccc|ccc|ccc||c@{}}
			\toprule
			\multicolumn{2}{@{}l@{}}{} &
			\multicolumn{3}{c}{\textbf{Tag}} &
			\multicolumn{3}{c}{\textbf{Spread}} &
			\multicolumn{3}{c}{\textbf{World}} &
			\multicolumn{1}{c@{}}{} \\
			
			\cmidrule(lr){3-5} \cmidrule(lr){6-8} \cmidrule(lr){9-11}

				\textbf{Alg} & \textbf{Dataset} & Medium & MD-Replay & Random & Medium & MD-Replay & Random & Medium & MD-Replay & Random & Average \\
				
				\midrule
				\multirow{6}{*}{CFCQL}
				& Original    & 44.3{\scriptsize$\pm$16.5} & 43.3{\scriptsize$\pm$13.7} &  7.4{\scriptsize$\pm$2.1} & 16.6{\scriptsize$\pm$9.8} & 31.2{\scriptsize$\pm$14.0} & 18.8{\scriptsize$\pm$7.6} & 82.0{\scriptsize$\pm$25.5} & 49.1{\scriptsize$\pm$8.5} &  7.9{\scriptsize$\pm$2.3} & 33.4 \\
				
                & MADiff*    & \textbf{57.8{\scriptsize$\pm$16.4}} & \underline{54.3{\scriptsize$\pm$5.8}} & \underline{12.2{\scriptsize$\pm$8.8}} &  19.6{\scriptsize$\pm$10.1} & \underline{58.8{\scriptsize$\pm$7.7}} & 20.7{\scriptsize$\pm$10.6} & 94.2{\scriptsize$\pm$9.3} & \underline{54.1{\scriptsize$\pm$9.4}} &  \underline{8.6{\scriptsize$\pm$3.3}} & \underline{42.3} \\
                
                & DOF*    & 35.0{\scriptsize$\pm$8.2} & 29.9{\scriptsize$\pm$4.9} & 9.3{\scriptsize$\pm$9.2} &  12.5{\scriptsize$\pm$6.7} & 54.1{\scriptsize$\pm$14.5} & 15.7{\scriptsize$\pm$6.6} & 88.8{\scriptsize$\pm$21.4} & 23.1{\scriptsize$\pm$8.0} &  3.4{\scriptsize$\pm$4.8} & 30.2 \\
				
                & MADiTS  & 52.8{\scriptsize$\pm$14.9} & 24.5{\scriptsize$\pm$6.0} &  6.7{\scriptsize$\pm$3.0} & \textbf{25.8{\scriptsize$\pm$8.9}} & 46.7{\scriptsize$\pm$12.3} & 21.6{\scriptsize$\pm$10.5} & 90.8{\scriptsize$\pm$22.3} & \textbf{58.6{\scriptsize$\pm$5.7}} &  3.7{\scriptsize$\pm$1.2} & 36.8 \\
                
                & INS  & 46.6{\scriptsize$\pm$7.8} & 45.8{\scriptsize$\pm$8.0} &  8.1{\scriptsize$\pm$5.4} & 18.5{\scriptsize$\pm$8.9} & 45.1{\scriptsize$\pm$14.4} & \textbf{25.8{\scriptsize$\pm$6.7}} & \underline{103.8{\scriptsize$\pm$19.1}} & 51.9{\scriptsize$\pm$10.4} &  8.1{\scriptsize$\pm$2.8} & 39.3 \\
     
                & Ours & \underline{54.7{\scriptsize$\pm$17.7}} & \textbf{54.4{\scriptsize$\pm$6.0}} & \textbf{13.3{\scriptsize$\pm$3.5}} & \underline{20.6{\scriptsize$\pm$9.5}} & \textbf{59.7{\scriptsize$\pm$16.6}} & \underline{25.2{\scriptsize$\pm$8.2}} & \textbf{111.4{\scriptsize$\pm$10.8}} &
                51.5{\scriptsize$\pm$8.0} & \textbf{10.0{\scriptsize$\pm$3.0}} & \textbf{44.5} \\

                \midrule
				\multirow{6}{*}{OMAR}
				& Original    & 10.8{\scriptsize$\pm$15.1} & 53.3{\scriptsize$\pm$12.1} & 21.7{\scriptsize$\pm$11.1} & -1.1{\scriptsize$\pm$6.4} & 32.8{\scriptsize$\pm$7.3} & 21.1{\scriptsize$\pm$7.0} & 63.5{\scriptsize$\pm$34.8} & 55.3{\scriptsize$\pm$8.9} &  7.1{\scriptsize$\pm$2.3} & 29.4 \\
				
                & MADiff*    & 18.4{\scriptsize$\pm$25.0} & \textbf{62.8{\scriptsize$\pm$8.1}} & 17.9{\scriptsize$\pm$5.3} & -1.0{\scriptsize$\pm$6.9} & \textbf{50.0{\scriptsize$\pm$10.1}} & 18.8{\scriptsize$\pm$7.6} & 58.0{\scriptsize$\pm$39.5} & \underline{59.1{\scriptsize$\pm$18.0}} &  \textbf{11.7{\scriptsize$\pm$3.4}} & \underline{32.9} \\
                
                & DoF*    & 7.1{\scriptsize$\pm$7.7} & 41.1{\scriptsize$\pm$7.0} & \underline{24.0{\scriptsize$\pm$7.2}} & -2.2{\scriptsize$\pm$13.7} & 39.4{\scriptsize$\pm$8.7} & 18.6{\scriptsize$\pm$8.0} & 68.9{\scriptsize$\pm$26.6} & 40.9{\scriptsize$\pm$7.5} &  \underline{10.4{\scriptsize$\pm$2.6}} & 27.6 \\
				
                & MADiTS  & \underline{23.9{\scriptsize$\pm$9.5}} & 41.5{\scriptsize$\pm$11.3} & \textbf{28.8{\scriptsize$\pm$10.8}} &  \underline{3.3{\scriptsize$\pm$7.6}} & 43.7{\scriptsize$\pm$12.6} & \underline{23.5{\scriptsize$\pm$11.3}} & 53.1{\scriptsize$\pm$19.3} & 56.1{\scriptsize$\pm$3.8} &  6.3{\scriptsize$\pm$2.3} & 31.2 \\
                
                & INS    & 4.5{\scriptsize$\pm$5.6} & 50.9{\scriptsize$\pm$8.5} & 18.1{\scriptsize$\pm$5.5} & 0.6{\scriptsize$\pm$8.9} & 45.4{\scriptsize$\pm$10.5} & 21.8{\scriptsize$\pm$5.1} & \underline{75.8{\scriptsize$\pm$29.3}} & 54.4{\scriptsize$\pm$11.5} &  7.8{\scriptsize$\pm$3.6} & 31.0 \\
                
                & Ours & \textbf{24.3{\scriptsize$\pm$21.7}} & \underline{60.9{\scriptsize$\pm$6.9}} & 19.1{\scriptsize$\pm$7.2} & \textbf{9.7{\scriptsize$\pm$4.8}} & \underline{46.5{\scriptsize$\pm$20.7}} & \textbf{27.0{\scriptsize$\pm$10.2}} & \textbf{109.7{\scriptsize$\pm$12.4}} &
                \textbf{59.5{\scriptsize$\pm$12.3}} & 10.3{\scriptsize$\pm$5.0} & \textbf{40.8} \\

                \midrule

				\multirow{6}{*}{MATD3+BC}
				& Original    & 23.4{\scriptsize$\pm$11.3} & 32.2{\scriptsize$\pm$9.0} &  3.7{\scriptsize$\pm$3.7} &  8.5{\scriptsize$\pm$7.3} & 28.1{\scriptsize$\pm$10.6} & 20.1{\scriptsize$\pm$7.2} & \underline{84.4{\scriptsize$\pm$31.3}} & 50.6{\scriptsize$\pm$8.9} &  4.4{\scriptsize$\pm$2.3} & 28.4 \\
				& MADiff*    & \underline{37.2{\scriptsize$\pm$27.3}} & \underline{49.2{\scriptsize$\pm$9.4}} &  \textbf{13.2{\scriptsize$\pm$9.2}} &  7.5{\scriptsize$\pm$6.4} & \underline{53.8{\scriptsize$\pm$18.7}} & 15.8{\scriptsize$\pm$8.1} & 82.5{\scriptsize$\pm$16.7} & 47.4{\scriptsize$\pm$10.6} &  7.1{\scriptsize$\pm$2.5} & \underline{34.9} \\
                
                & DoF*    & 22.4{\scriptsize$\pm$9.8} & 32.8{\scriptsize$\pm$7.6} &  3.4{\scriptsize$\pm$1.3} &  \underline{9.6{\scriptsize$\pm$6.7}} & 49.4{\scriptsize$\pm$18.6} & 19.1{\scriptsize$\pm$8.9} & 64.5{\scriptsize$\pm$33.0} & 26.2{\scriptsize$\pm$8.5} &  \textbf{9.8{\scriptsize$\pm$6.0}} & 26.4 \\
				
                & MADiTS  & \textbf{39.9{\scriptsize$\pm$6.2}} & 22.0{\scriptsize$\pm$4.6} &  2.5{\scriptsize$\pm$1.2} & \textbf{11.0{\scriptsize$\pm$3.3}} & 45.3{\scriptsize$\pm$10.5} & 17.4{\scriptsize$\pm$9.0} & 75.0{\scriptsize$\pm$12.3} & \textbf{60.7{\scriptsize$\pm$8.8}} &  1.5{\scriptsize$\pm$2.5} & 30.6 \\
                
                & INS    & 15.7{\scriptsize$\pm$9.7} & 43.9{\scriptsize$\pm$7.0} &  4.6{\scriptsize$\pm$3.7} &  5.6{\scriptsize$\pm$15.9} & 45.9{\scriptsize$\pm$20.7} & \underline{21.6{\scriptsize$\pm$8.2}} & 66.5{\scriptsize$\pm$34.6} & 49.0{\scriptsize$\pm$12.4} &  8.0{\scriptsize$\pm$1.8} & 29.0 \\
				
                
                & Ours & 32.8{\scriptsize$\pm$21.6} & \textbf{50.8{\scriptsize$\pm$8.0}} & \underline{10.9{\scriptsize$\pm$4.6}} & 7.8{\scriptsize$\pm$5.1} & \textbf{54.2{\scriptsize$\pm$17.9}} & \textbf{24.7{\scriptsize$\pm$10.1}} & \textbf{100.2{\scriptsize$\pm$8.8}} &
                \underline{52.2{\scriptsize$\pm$10.3}} & \underline{8.2{\scriptsize$\pm$4.1}} & \textbf{38.0} \\
                
				\bottomrule
			\end{tabular}
    }
\end{table*}
\begin{table*}[t]
    \vspace{-0.8em}
	\small
    \caption{Test win rate in the SMAC environment (\textbf{bold}: best, \underline{underline}: second best)}
    \vspace{-0.4em}
    \label{table:smac_main_result}
	\centering
    \renewcommand{\arraystretch}{1.15}
    \robustify\bfseries
    \resizebox{0.9\textwidth}{!}{%
      \setlength{\tabcolsep}{4pt}\begin{tabular}{@{}l||l|cc|cc|cc||c@{}}
			\toprule		
			\multicolumn{2}{@{}l@{}}{} &
			\multicolumn{2}{c}{\textbf{3m}} &
			\multicolumn{2}{c}{\textbf{2s3z}} &
			\multicolumn{2}{c}{\textbf{8m}} &
			\multicolumn{1}{c@{}}{} \\
			\cmidrule(lr){3-4}\cmidrule(lr){5-6}\cmidrule(lr){7-8}
			
			\textbf{Alg} & \textbf{Dataset} & Medium & Good & Medium & Good & Medium & Good & Average \\
			
			\midrule
			\multirow{6}{*}{CFCQL}
			& Original & 0.75{\scriptsize$\pm$0.02}  & \underline{0.96{\scriptsize$\pm$0.04}} &  \underline{0.38{\scriptsize$\pm$0.38}} &  0.40{\scriptsize$\pm$0.55}  & 0.57{\scriptsize$\pm$0.52}  & 0.80{\scriptsize$\pm$0.45}  & 0.64 \\
			& MADiff* & 0.78{\scriptsize$\pm$0.27}  & 0.84{\scriptsize$\pm$0.36} &  0.31{\scriptsize$\pm$0.18} &  0.68{\scriptsize$\pm$0.23}  & 0.80{\scriptsize$\pm$0.45}  & 0.87{\scriptsize$\pm$0.13}  & \underline{0.71} \\
                & DoF* & \underline{0.89{\scriptsize$\pm$0.14}}  & 0.86{\scriptsize$\pm$0.09} &  0.16{\scriptsize$\pm$0.49} &  0.76{\scriptsize$\pm$0.33}  & 0.78{\scriptsize$\pm$0.25}  & 0.73{\scriptsize$\pm$0.19}  & 0.70 \\
			& MADiTS & 0.70{\scriptsize$\pm$0.14}  & 0.86{\scriptsize$\pm$0.27} &  0.30{\scriptsize$\pm$0.65} &  0.51{\scriptsize$\pm$0.14}  & \textbf{0.90{\scriptsize$\pm$0.08}}  & \underline{0.89{\scriptsize$\pm$0.12}}  & 0.69 \\
                & INS & 0.82{\scriptsize$\pm$0.31}  & 0.79{\scriptsize$\pm$0.46} &  0.17{\scriptsize$\pm$0.7} &  \textbf{0.83{\scriptsize$\pm$0.21}}  & 0.82{\scriptsize$\pm$0.33}  & 0.85{\scriptsize$\pm$0.16}  & \underline{0.71} \\
            & Ours &\textbf{1.00{\scriptsize$\pm$0.00}}  & \textbf{0.97{\scriptsize$\pm$0.07}} &  \textbf{0.43{\scriptsize$\pm$0.58}} &  \underline{0.77{\scriptsize$\pm$0.09}}  & \underline{0.84{\scriptsize$\pm$0.06}}  & \textbf{1.00{\scriptsize$\pm$0.00}}  & \textbf{0.84} \\
			\midrule
			\multirow{6}{*}{OMAR}
			& Original & 0.71{\scriptsize$\pm$0.07}  & 0.41{\scriptsize$\pm$0.24} &  \textbf{0.40{\scriptsize$\pm$0.55}} &  0.60{\scriptsize$\pm$0.55}  & 0.75{\scriptsize$\pm$0.43}  & 0.75{\scriptsize$\pm$0.43}  & 0.60 \\
			& MADiff* & 0.71{\scriptsize$\pm$0.30}  & 0.53{\scriptsize$\pm$0.42} &  0.37{\scriptsize$\pm$0.11} &  0.53{\scriptsize$\pm$0.19}  & 0.89{\scriptsize$\pm$0.11}  & 0.69{\scriptsize$\pm$0.38}  & 0.62 \\
                & DoF* & 0.82{\scriptsize$\pm$0.18}  & \underline{0.91{\scriptsize$\pm$0.14}} &  \textbf{0.40{\scriptsize$\pm$0.42}} &  0.76{\scriptsize$\pm$0.19}  & 0.87{\scriptsize$\pm$0.07}  & 0.74{\scriptsize$\pm$0.28}  & \underline{0.75} \\
			& MADiTS & 0.72{\scriptsize$\pm$0.12}  & 0.44{\scriptsize$\pm$0.40} &  0.36{\scriptsize$\pm$0.60} &  0.40{\scriptsize$\pm$0.29}  & 0.72{\scriptsize$\pm$0.26}  & 0.89{\scriptsize$\pm$0.03}  & 0.59 \\
                & INS & \underline{0.84{\scriptsize$\pm$0.20}}  & 0.88{\scriptsize$\pm$0.14} &  0.36{\scriptsize$\pm$0.37} &  \underline{0.78{\scriptsize$\pm$0.35}}  & 0.85{\scriptsize$\pm$0.14}  & 0.81{\scriptsize$\pm$0.15}  & \underline{0.75} \\
            & Ours & \textbf{1.00{\scriptsize$\pm$0.00}}  & \textbf{0.99{\scriptsize$\pm$0.02}} &  0.37{\scriptsize$\pm$0.58} & \textbf{0.94{\scriptsize$\pm$0.00}}  & 0.83{\scriptsize$\pm$0.02}  & \textbf{1.00{\scriptsize$\pm$0.00}}  & \textbf{0.86} \\
			\bottomrule
		\end{tabular}
    }
    \vspace{-1.2em}
\end{table*}

The generation order affects how much flexibility remains for coordinating agents during sequential generation. If an agent that strongly constrains later trajectory evolution is generated too early, subsequent agents may have limited flexibility to coordinate with it. MASTARS therefore uses subgoal timesteps to choose the order $(\kappa(1), \ldots, \kappa(N))$. Since an agent with an earlier subgoal has a longer suffix to regenerate, generating it later allows the agent to better adapt to the trajectories generated before it.

\textbf{Order of Generation.}~~
The subgoal $g^i$ of agent $i$ is the timestep at which its estimated value is maximized. We sort agents in descending order of their subgoal indices:
\begin{align}\label{eq:order_generation}
\small
(\kappa(1), \ldots, \kappa(N))
= \mathrm{argsort}\bigl[g^1, \ldots, g^N\bigr]_{\downarrow},
\end{align}
where $\downarrow$ denotes ``sorting in descending order''.
This ordering generates agents with later subgoals first and defers agents with earlier subgoals to later stages, preserving flexibility for collision avoidance and multi-agent coordination.

Note that preserving the relative ordering of state values is important for subgoal identification and generation ordering in MASTARS. Fortunately, since values are needed only for observed states, SARSA-style learning in Eq.~\eqref{eq:value_loss} in Appendix \ref{appendix:training_loss} is sufficient, avoiding extrapolation and improving stability.

\paragraph{Practical Implementation.}
MASTARS leverages five pre-trained models: a diffusion model $\epsilon_\theta$ for generating observation sequences (built on the MADiff~\citep{Zhu024madiff} architecture), three dynamics models (a reward function $g_\xi$, an inverse dynamics model $I_\phi$, and a transition model $f_\varphi$) that produce the corresponding actions and rewards from a generated observation sequence, and a value function $V_\zeta$. Their training losses are given by Eqs.~\eqref{eq:rg_diff_loss},~\eqref{eq:dynamic_model_loss}, and~\eqref{eq:value_loss}, with full details in Appendix~\ref{appendix:training_loss}. The overall pipeline of MASTARS is illustrated in Fig.~\ref{fig:overflow_v1}.

In summary, MASTARS is a plug-in diffusion-based framework for multi-agent trajectory augmentation that integrates (i) agent-wise sequential generation with inpainting, (ii) subgoal-based partial generation, and (iii) subgoal-based ordering. By combining all three, MASTARS produces diverse, coordinated, and globally coherent trajectories. The complete algorithm is provided in Appendix~\ref{appendix:pseudocode}.

\vspace{-.5em}
\section{Experiments}\label{experiments}
\vspace{-.5em}
Our evaluation focuses on five aspects: (i) performance gains with the augmented dataset, (ii) visual analysis of our generation strategy, (iii) quantitative analysis of the generated dataset in terms of diversity, coordination, and coherence, (iv) reliability of the auxiliary models, and (v) the contribution of key components through an ablation study.

\subsection{Experimental Setup}\label{exp:experiment_setup}
\begin{figure*}[t]
    \centering
    \includegraphics[width=\linewidth]{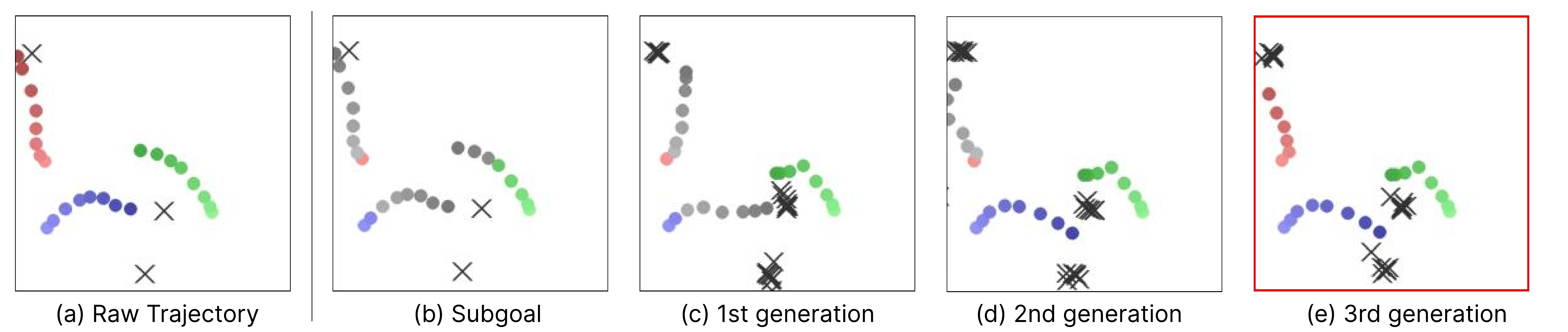}
    \caption{
    \textbf{Visualization of the full trajectory generation process.} Agents are generated sequentially via subgoal-conditioned inpainting, with each panel showing the trajectory after one agent is added, from the raw input (left) to the final result (right). \texttt{X} marks are landmark positions reconstructed from each agent's observation.
    }
    \label{fig:trajectory_viz}
    \vspace{-1em}
\end{figure*}
We evaluated MASTARS using three widely adopted offline MARL algorithms: CFCQL \citep{Shao23cfcql}, OMAR \citep{Pan22omar}, and MA-TD3+BC \citep{Fujimoto21td3bc}. These algorithms were tested on the MPE \citep{Lowe17mpe}, SMAC \citep{Samvelyan19smac}, and SMAC-v2 \citep{ellis2023smacv2} benchmarks. Detailed information regarding the benchmarks, algorithm configurations, and dataset compositions is provided in Appendix \ref{appendix:experiment_detail}.

\textbf{Baselines}~~We compared our method against other data augmentation approaches\footnote{An asterisk (*) indicates methods that are not trajectory generators; we use only their generation components.}: 
1) \textbf{Original}: using only the given offline dataset. 
2) \textbf{MADiff*} \citep{Zhu024madiff}: a diffusion-based trajectory generator with inter-agent attention. 
3) \textbf{DoF*} \citep{li2025dof}: a diffusion-based MARL policy extending the IGM principle to an IGD formulation.
4) \textbf{MADiTS} \citep{yuan25madits}: a stitching approach that constructs trajectories by combining segments.
5) \textbf{INS} \citep{fuins}: a sparse-attention method that explicitly models inter-agent interactions.

\subsection{Main Results on Offline MARL Benchmarks}
\textbf{Evaluation on MPE \& SMAC benchmarks}~
Tables~\ref{table:mpe_main_result} \& \ref{table:smac_main_result} demonstrate strong performance across diverse benchmarks and offline MARL algorithms. In the MPE benchmark, MASTARS enhances the average performance of CFCQL, OMAR, and MATD3+BC by 33.2\%, 38.8\%, and 33.8\%, respectively, compared to the original datasets. Notably, it achieves top-two results in 22 of 27 tasks.

Building on these results, the superiority of our framework extends to the SMAC benchmark, which presents more complex coordination challenges. Here, MASTARS improves CFCQL and OMAR by 31.3\% and 43.3\%, respectively, achieving top-two results in 10 out of 12 scenarios. Most notably, it achieves a perfect 100\% win rate in demanding tasks such as 3m (Medium) and 8m (Good).

\begin{wraptable}[9]{r}{0.5\textwidth}
    \centering
    \vspace{-1.3em}
    \small
    \caption{Average test win rate in SMAC-v2}
    \label{table:smacv2_summary_result}
    \resizebox{\linewidth}{!}{%
      \begin{tabular}{l||ccc||c}
    \toprule
    \textbf{Dataset} & \textbf{Protoss} & \textbf{Terran} & \textbf{Zerg} & \textbf{Average} \\
    \midrule
    Original & 0.12 & 0.22 & \underline{0.24} & 0.19 \\
    MADiff* & 0.15 & 0.30 & 0.14 & 0.21 \\
    DoF* & \underline{0.23} & 0.29 & 0.19 & \underline{0.25} \\
    MADiTS   & 0.18 & 0.18 & 0.19 & 0.18 \\
    INS      & 0.21 & \underline{0.33} & 0.15 & \underline{0.25} \\
    Ours     & \textbf{0.50} & \textbf{0.50} & \textbf{0.40} & \textbf{0.48} \\
    \bottomrule
\end{tabular}
    }
\end{wraptable}
\textbf{Extended evaluation on SMAC-v2 benchmark}~
To further evaluate the robustness of our framework in more challenging settings, we conducted extended experiments on the SMAC-v2 benchmark, which is characterized by significantly higher stochasticity and complexity. 
As summarized in Table~\ref{table:smacv2_summary_result}, MASTARS demonstrates remarkable adaptability to these environments, achieving a substantial average performance boost of 152.6\% compared to the original dataset. It reliably ranks within the top-two in 18 out of 20 tasks. Detailed environment descriptions, dataset composition, and full SMAC-v2 results are provided in Appendix \ref{appendix:smacv2}.

\vspace{-1em}
\paragraph{Summary of overall performance}
These results show that MASTARS consistently outperforms the evaluated augmentation baselines across environments, datasets, and offline MARL algorithms. Implementation details and computational cost analyses are provided in Appendices \ref{appendix:implementation_detail} and \ref{appendix:computational_cost}.

\subsection{Visualization Analysis of Our Generation Strategy}
\begin{figure}[t]
    \centering
    \includegraphics[width=0.78\linewidth]{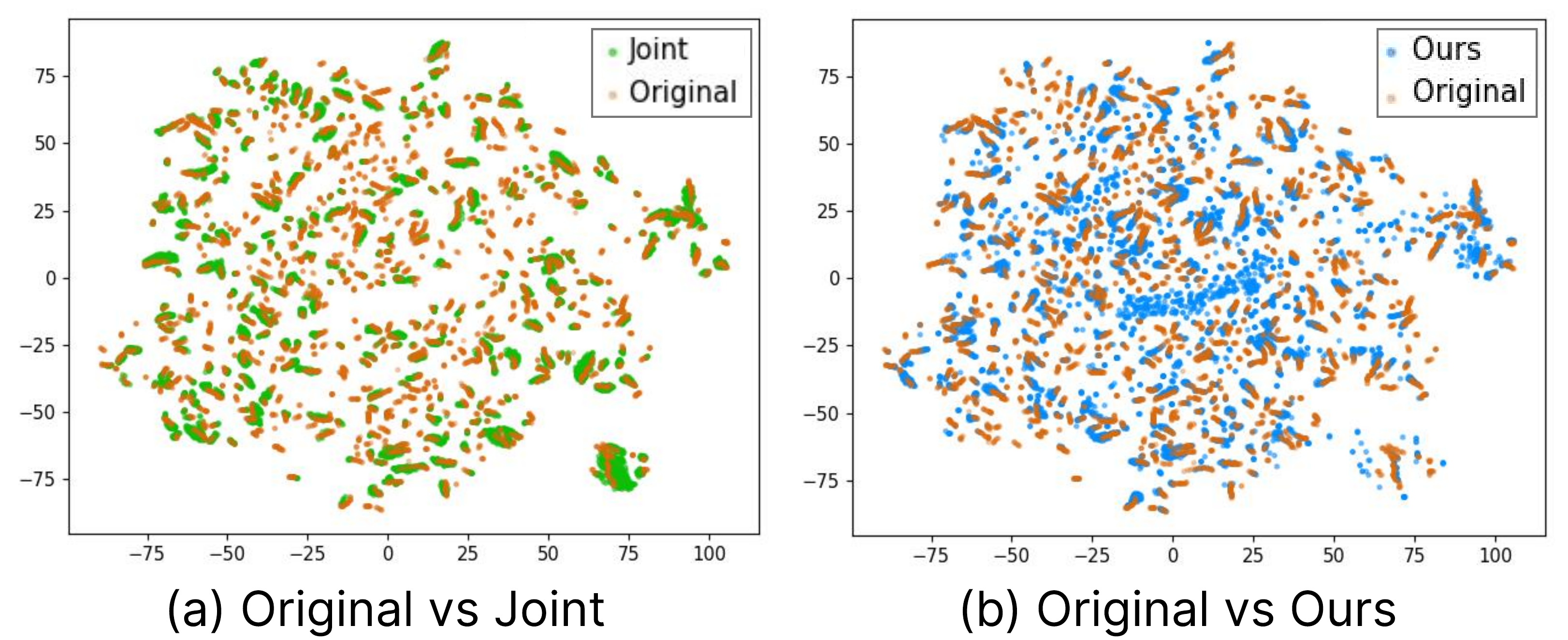}
    \vspace{-0.5em}
    \caption{t-SNE comparison: original (orange), jointly generated (green), and ours (blue)}
    \label{fig:tsne}
    \vspace{-1em}
\end{figure}

\paragraph{Sequential Trajectory Generation.}
We visualize a trajectory generated by MASTARS to illustrate its sequential construction process. As shown in Fig.~\ref{fig:trajectory_viz}, agents generate trajectories in the order of their subgoal timesteps in the original trajectory: green, blue, and then red. The green agent, with the latest subgoal, generates first, followed by the blue, and finally the red agent with the earliest subgoal.

By reusing trajectory segments up to the subgoal, MASTARS avoids full trajectory generation while leveraging expert behavior. Moreover, since agents with earlier subgoals offer greater flexibility, generating them later enhances overall diversity. As a result, MASTARS produces realistic and cooperative trajectories that avoid collisions and reach targets (more examples in Appendix~\ref{appendix:visual_mpe_all}).

\begin{wrapfigure}[8]{r}{0.45\linewidth}
    \centering
    \vspace{-1.4em}
    \includegraphics[width=\linewidth]{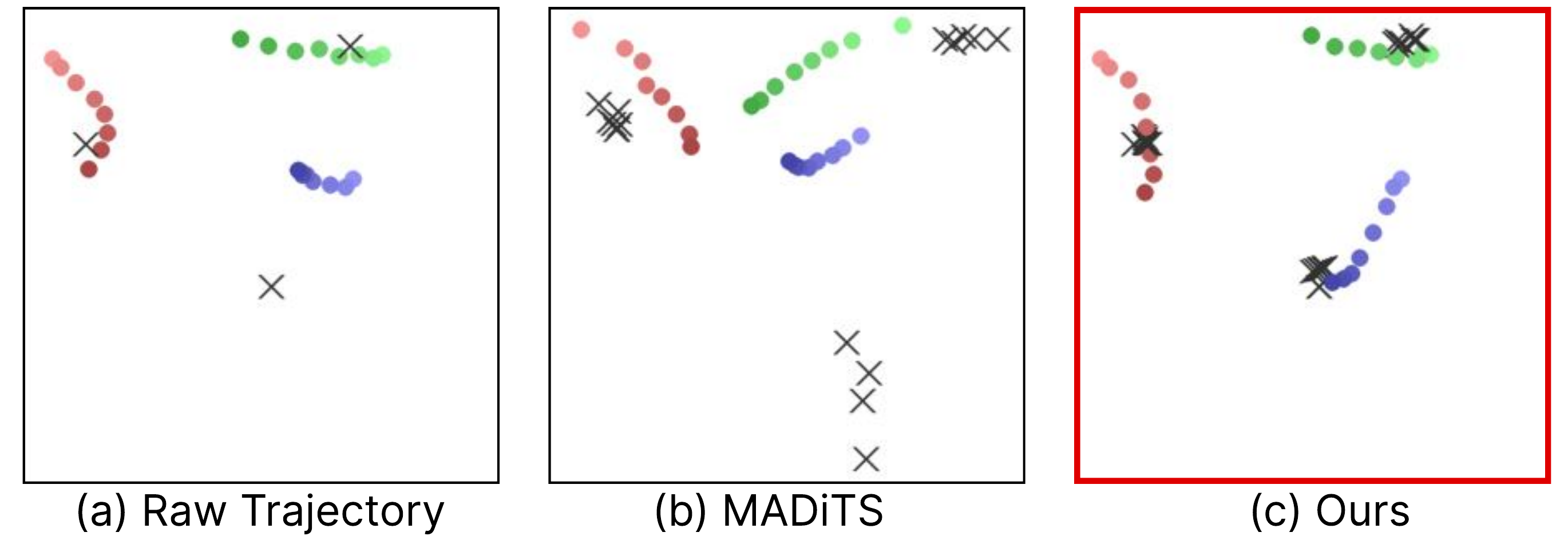}
    \vspace{-1.2em}
    \caption{\small Comparison of MADiTS and Ours}
    \label{fig:madits}
\end{wrapfigure}
\paragraph{Comparison with Multi-Agent Baselines.}
We visually compare MASTARS with MADiTS in Fig.~\ref{fig:madits}; INS generates transitions rather than trajectories and is thus not visualized. MADiTS exhibits incoherent and uncoordinated movements due to its one-shot refinement, whereas MASTARS produces coherent and well-coordinated trajectories. Additional comparisons are provided in Appendix~\ref{appendix:visual_madits}.


\subsection{Quantitative Analysis of Generated Dataset}\label{exp:quality}

\paragraph{Diversity.}
To verify the diversity of trajectories generated by MASTARS, we conduct a dataset analysis using DIRE and t-SNE. First, we adopt the DIRE metric \citep{wang2023dire}, which was proposed to distinguish between original and generated samples.
This metric is defined as:
$DIRE(x_0) = \|x_0 - R(I(x_0))\|$,
where $I(\cdot)$ and $R(\cdot)$ denote the forward and deterministic reverse processes, respectively.
Since $R(I(x_0))$ represents the reconstruction of $x_0$, DIRE measures the reconstruction error with respect to the training distribution.
Lower scores indicate closer alignment with the training data, and higher scores reflect greater distributional diversity of the generated trajectories relative to the training set.

\begin{wraptable}[5]{r}{0.4\linewidth}
    \centering
    \vspace{-1.2em}
    \small
    \caption{Reconstruction error (DIRE)}
    \label{tab:dire}
    \vspace{-0.6em}
    \begin{adjustbox}{width=\linewidth,center}
    \robustify\bfseries
    \begin{tabular}{l|ccc}
        \toprule
        ~ & Original & MADiff* & Ours \\
        \midrule
        DIRE & 0.045 & 0.086 & \textbf{0.101} \\
        \bottomrule
    \end{tabular}
    \end{adjustbox}
\end{wraptable}
As shown in Table~\ref{tab:dire}, MASTARS yields higher DIRE scores than joint generation methods, such as MADiff*, indicating the synthesis of more novel behaviors.
This observation is further corroborated by the t-SNE visualization in Fig.~\ref{fig:tsne}.
While MADiff* overlaps significantly with the original data manifold, MASTARS produces a broader distribution that extends outside the original data manifold. Note that all analyzed samples have passed return-conditioned guidance (Eq.~\eqref{eq:rg_diff_loss}) and transition validation (Eq.~\eqref{eq:sample_transition}), so this broader coverage reflects guided, filtered samples rather than raw generations.

\paragraph{Coordination and Coherence.}
To quantify the visual analyses in Figs.~\ref{fig:motivation} and \ref{fig:madits}, we measure three metrics over all generated MPE trajectories: \emph{collision rate}, the fraction of transitions with an agent collision (coordination); \emph{landmark error}, the within-episode dispersion of each fixed landmark's position reconstructed from every agent's generated observation, i.e., the mean $\ell_2$ distance of the reconstructions to their centroid, which is zero for real data (coherence); and \emph{dynamics violation rate}, the fraction of generated transitions that deviate from the true environment dynamics by more than the threshold $\epsilon$ in Eq.~\eqref{eq:sample_transition}.
As shown in Table~\ref{tab:quality}, MASTARS achieves the best result on all three metrics: sequential inpainting improves coordination and coherence, while transition validation removes inconsistent candidates. Together, these results show that MASTARS synthesizes diverse, coordinated, and globally coherent trajectories.

\begin{table}[!h]
    \centering
    \small
    \caption{Coordination and coherence metrics of generated MPE datasets. INS generates transitions, so its landmark error is undefined. Baselines use looser acceptance thresholds to retain sufficient data.}
    \label{tab:quality}
    \vspace{-0.5em}
        \begin{adjustbox}{width=0.8\linewidth,center}
    \robustify\bfseries
    \begin{tabular}{l|ccc}
        \toprule
        Method & Collision rate $\downarrow$ & Landmark error $\downarrow$ & Dynamics violation $\downarrow$ \\
        \midrule
        MADiff* & 13.3\% & 0.088 & 14.2\% \\
        DoF*    & 42.0\% & 0.276 & 85.9\% \\
        MADiTS  & 33.2\% & 0.078 & 19.0\% \\
        INS     & 16.7\% & N/A   & 93.0\% \\
        \midrule
        MASTARS (Ours) & \textbf{8.4\%} & \textbf{0.039} & \textbf{2.0\%} \\
        \bottomrule
    \end{tabular}
    \end{adjustbox}
    \vspace{-1.6em}
\end{table}

\begin{table}[!ht]
    \centering
    \small
    \caption{Prediction errors of auxiliary models on offline and generated data. N/A: no ground-truth actions or values exist for generated data.}
    \label{tab:auxiliary}
    \vspace{-0.6em}
    \begin{adjustbox}{width=\linewidth,center}
    \robustify\bfseries
    \begin{tabular}{l|cc||l|cc}
        \toprule
        Model & Offline & Generated & Model & Offline & Generated \\
        \midrule
        Reward $g_\xi$            & 2.3\% (1.7\%) & 4.0\% (2.3\%) & Transition $f_\varphi$     & 0.9\% (0.6\%) & 0.8\% (0.6\%) \\
        Inverse dynamics $I_\phi$ & 5.9\% (4.3\%) & N/A           & Value $V_\zeta$ (TD error) & 0.01          & N/A \\
        \bottomrule
    \end{tabular}
    \end{adjustbox}
    \vspace{-1.4em}
\end{table}


\newpage
\subsection{Reliability of Auxiliary Models}\label{exp:auxiliary}

MASTARS relies on four auxiliary models ($g_\xi$, $I_\phi$, $f_\varphi$, $V_\zeta$), raising the concern that their errors may compound. However, none of them is used for recursive rollouts: $V_\zeta$ only selects subgoals and the order, $f_\varphi$ only validates candidates, and $g_\xi$ and $I_\phi$ only annotate generated observations, so their errors do not accumulate across generation steps.

Table~\ref{tab:auxiliary} shows that all prediction errors remain within a few percent of the data range, even on generated data. MASTARS is also robust to these errors: $V_\zeta$ uses only the relative ranking of values, so value noise of up to 20\% still outperforms Original (Appendix~\ref{appendix:value_sensitivity}); $f_\varphi$ affects MASTARS only through $\epsilon$, to which performance is robust (Fig.~\ref{fig:hyperparam_sensitivity}); and reward and action noise of up to 30\% keeps the score at 37.4--40.3 (Appendix~\ref{appendix:annotation_noise}). Thus, MASTARS is robust to auxiliary-model errors.

\subsection{{Ablation Studies}}

We conducted ablation studies to assess the contributions of four key components of MASTARS: (1) partial generation via subgoal conditioning, (2) sequential generation, (3) inpainting, and (4) subgoal-based ordering. All ablation results are averaged over CFCQL, OMAR, and MATD3+BC on the Medium, MD-Replay, and Random datasets of each MPE task. 

\begin{wraptable}[9]{r}{0.52\textwidth}
    \centering
    \vspace{-1.4em}
    \small
    \caption{Ablation performance}
    \vspace{-.5em}
    \label{tab:ablation_key_component}
    \begin{adjustbox}{width=\linewidth,center}
        \robustify\bfseries
        \begin{tabular}{l|ccc|c}
        \toprule
        ~ & Tag & Spread & World & Total \\
        \midrule
        w/o Subgoal & 28.11 & 26.06 & 45.84 & 33.34 \\
        w/o Sequence & 27.35 & 23.88 & 52.58 & 34.60 \\
        w/o Inpainting  & 29.00 & 27.96 & 54.86 & 37.27 \\
        \midrule
        Random ordering & 28.03 & 24.11 & 46.46 & 32.87 \\
        Reverse ordering & 27.60 & 24.31 & 53.73 & 35.21 \\
        \midrule
        Ours & \textbf{35.69} & \textbf{30.60} & \textbf{57.00} & \textbf{41.10}\\
        \bottomrule
        \end{tabular}
    \end{adjustbox}
    \vspace{-1em} 
\end{wraptable} 

Since the ordering mechanism is foundational, we tested its necessity by replacing our subgoal-based strategy with reverse and random ordering schemes. As shown in Table~\ref{tab:ablation_key_component}, removing individual modules (``w/o") or using alternative ordering schemes leads to significant performance degradation.

Additional analyses of hyperparameter sensitivity, synthesis fraction, scalability with respect to agent number and horizon length, and value-estimation sensitivity are provided in Appendix~\ref{appendix:additional_experiments}, and generation cost in Appendix~\ref{appendix:computational_cost}.

\section{Conclusion}

In this paper, we showed that effective offline MARL augmentation requires more than realistic joint trajectory modeling.  To overcome the limitation of conventional data augmentation methods, we introduced MASTARS, a novel data augmentation method for offline MARL. 
By decomposing generation agent-wise while using inpainting to preserve joint coherence, and by preserving high-value subgoal prefixes, 
MASTARS expands the offline dataset in directions that remain coordinated and globally feasible. This combination leads to consistent improvements across multiple offline MARL algorithms and benchmarks, suggesting that structured generative augmentation is a promising direction for improving data-limited multi-agent learning.

\section*{Acknowledgments}
This work was supported by the National Research Foundation of Korea (NRF) grant funded by the Korea government (MSIT) (No.\ RS-2025-00557589, Generative Model Based Efficient Reinforcement Learning Algorithms for Multi-modal Expansion in Generalized Environments), the National Research Foundation of Korea (NRF) grant funded by the Korea government (MSIT) (No.\ RS-2026-25510069), and the ``Advanced GPU Utilization Support Program'' funded by the Government of the Republic of Korea (Ministry of Science and ICT) (No.\ 02-26-01-0110). The authors declare no competing interests.

\bibliographystyle{plainnat}
\bibliography{neurips_2026}







\newpage
\appendix

\section{Broader Impact}
\label{appendix:broader_impact}
Our algorithm can generate cooperative trajectories from small datasets, reducing the need for expensive data collection in real-world applications. This has significant potential in domains where reinforcement learning is applicable but data acquisition is costly, such as autonomous driving, warehouse robotics, and surgical training simulators. By lowering the data requirements, our method enables faster prototyping and broader accessibility of RL-based systems. However, careful validation is necessary to ensure that the generated data does not introduce unrealistic behaviors or biases when deployed in safety-critical environments. 

\vspace{3em}
\section{Limitations and Future Works}
\label{appendix:limitation}
The sequential generation process in MASTARS performs one inpainting pass per agent, so its sampling cost grows with the number of agents. Although this cost is incurred only once when constructing the augmented dataset and is offset by a higher acceptance rate under strict validation, it may become a practical burden in settings with very many agents; reducing it, e.g., by generating weakly interacting agents in parallel, is a promising direction. In addition, it would be beneficial to explore alternative approaches to subgoal generation, such as learning subgoal distributions or leveraging a combination of joint and individual value functions \citep{jeon2022maser}.


\newpage
\section{Trajectory Generation in Single-Agent RL}
\label{appendix:single_agent_diffusion_model}

In offline RL, diffusion models are trained to generate entire trajectory segments by treating each segment as if it were a 2D “image.” Concretely, they learn the distribution
$
    p_\theta\bigl(\tau_{t:t+H-1}\bigr),
$
by minimizing the DDPM loss in Eq.~\eqref{eq:ddpm_loss}. To do so, each trajectory segment is reshaped into a matrix of size $(d_s + d_a)\times H$ by stacking state and action vectors as rows, where $d_s$ is the state dimension, $d_a$ is the action dimension, and $H$ is the segment length to be generated. The diffusion model then generates this matrix just as it would generate an image:
\begin{align}
\tau = 
\begin{bmatrix}
\mathbf{s}_t & \mathbf{s}_{t+1} & \cdots & \mathbf{s}_{t+H-1} \\
\mathbf{a}_t & \mathbf{a}_{t+1} & \cdots & \mathbf{a}_{t+H-1}
\end{bmatrix}.
\end{align}
By framing trajectories in this way, the model naturally outputs new trajectory “images” whose statistical properties match those of the offline dataset.

\vspace{3em}
\section{Trajectory Generation in MADiff}
\label{appendix:madiff_data_generation}

Given a sequence of observations $\tau = (\boldsymbol{o}_{t:t+H-1})$, MADiff \citep{Zhu024madiff} proposes two trajectory generation strategies: Independent Trajectory Generation and Joint Trajectory Generation. 
 
\paragraph{Independent Trajectory Generation.} In the independent scheme, we adopt a decentralized execution scheme in MADiff. Each agent $i$ generates its trajectory based solely on its current local observation, without access to the observations or plans of other agents. The diffusion process begins with a noisy trajectory conditioned on agent $i$'s current observation, represented as
{\footnotesize $\tilde{\boldsymbol{x}}_K = [[\tilde x^1_{K,t}, \ldots, \tilde x^1_{K,t_e}], \ldots, [o_{t_s}^i, \ldots, \tilde x^i_{K,t_e}], \dots, [\tilde x^N_{K,t}, \ldots, \tilde x^N_{K,t_e}]]$},
with {\footnotesize $\tilde{x}^j_{K,t'}{\sim}\mathcal{N}(0,1)$}, and {\footnotesize $t_e = t_s{+}H{-}1$}.

\paragraph{Joint Trajectory Generation.}
The joint generation scheme adopts a centralized control setup as in MADiff, where the trajectories of all agents are generated simultaneously, conditioned on the current joint observation $\boldsymbol{o}_t$. The initial noisy trajectory is constructed as
{\footnotesize $\tilde{\boldsymbol{x}}_K = [[ o^1_{t_s}, \ldots, \tilde x^1_{K,t_e}], \ldots, [o_{t_s}^i, \ldots, \tilde x^i_{K,t_e}], \dots, [o^N_{t_s}, \ldots, \tilde x^N_{K,t_e}]]$},
again with {\footnotesize$\tilde{x}^j_{K,t'}{\sim}\mathcal{N}(0,1)$}, and {\footnotesize $t_e{=}t_s{+}H{-}1$}.

\paragraph{Simple Sequence Trajectory Generation.} 
Since MADiff does not propose a sequential generation scheme, we propose a simple, naive sequential trajectory generation scheme as follows. In the simple sequence generation scheme, trajectories are generated sequentially following a predefined order of agents (we use the same order as our subgoal-based ordering to visualize our motivation). For each agent $i \in \{1, \dots, N\}$, the diffusion process is performed to generate its specific trajectory while keeping others fixed. The initial noisy trajectory $\tilde{\boldsymbol{x}}_K$ at step $i$ is constructed as follows:
{\footnotesize $\tilde{\boldsymbol{x}}_K = [[\hat{x}^1_{t}, \ldots, \hat{x}^1_{t_e}], \dots, [o_{t}^i, \dots, \tilde{x}^i_{K,t_e}], \dots, [\tilde{x}^N_{K,t}, \dots, \tilde{x}^N_{K,t_e}]]$},
where $\hat{x}^j$ for $j < i$ represents the fixed trajectory already generated in previous steps, and $\tilde{x}^j_{K,t'} \sim \mathcal{N}(0,1)$ for $j \geq i$ denotes Gaussian noise for the current and subsequent agents. This procedure is repeated until the entire sequence of agents is finalized, resulting in a complete joint trajectory.

\clearpage{}

\section{Pseudocode and Loss functions}\label{appendix:pseudocode_and_training_loss}
This section presents the algorithm in detail using pseudocode and outlines the loss functions of the pretrained models used for generation.

\subsection{Pseudocode}\label{appendix:pseudocode}

\begin{algorithm}[H] 
   \caption{Agent-Wise Sequential Trajectory Generation}
   \label{alg:Trajectory_generation}
\begin{algorithmic}[1]
   \STATE {\bfseries Input:} Original offline dataset $D_{\text{org}}$,
  Pretrained diffusion model $\epsilon_\theta$,
  Inverse dynamics model $I_\phi$,
  Transition model $f_\varphi$,
  Reward model $g_\xi$,
  Value function $V_\zeta$, 
  Number of episodes to generate $N_\text{episode}$,
  Number of denoising steps $K$,
  Pre-defined schedule $\bar{\alpha}_k$
  \STATE {\bfseries Output:} Augmented dataset $D_{\text{aug}}$
   \STATE Initialize $D_{\text{aug}} \leftarrow \emptyset$
   
   \WHILE{number of episodes in $D_{\text{aug}} < N_{\text{episode}}$}
      \STATE Sample joint sequence $\tau^0 = (\mathbf{o}_t, \dots, \mathbf{o}_{t+H-1}) \sim D_{\text{org}}$
      
      \STATE 
      \textcolor{blue!50}{// Subgoal-Based Ordering}
      \STATE Calculate $\text{Order}(\kappa(1), \dots, \kappa(N))$ from $\tau^0$ via Eq.~\eqref{eq:order_generation}
      \STATE $\texttt{Input\_traj} \leftarrow \tau^0$

      \STATE \textcolor{blue!50}{// Agent-Wise Sequential Generation}
      \FOR{each agent $i$ in $\text{Order}(\kappa(1), \dots, \kappa(N))$}
        \STATE Select $g^i$ as the observation in $\texttt{Input\_traj}$ with the highest value $V_\zeta$.
        \STATE Generate mask $m$ with $g^i$ via Eq.~\eqref{eq:mask_in_Repaint_with_subgoal}
        \STATE \textcolor{gray}{// Initialize at Step $K$}
        \STATE $\tilde{\tau}_K^{\text{posterior}} \sim \mathcal{N}(\mathbf{0}, \mathbf{I})$ 
        \STATE $\tilde{\tau}_K^{\text{prior}} \sim \mathcal{N}(\sqrt{\bar{\alpha}_K} \texttt{Input\_traj}, (1 - \bar{\alpha}_K)\mathbf{I})$
        \STATE $\tilde{\tau}_K \leftarrow m \odot \tilde{\tau}_K^{\text{posterior}} + (1-m) \odot \tilde{\tau}_K^{\text{prior}}$
         \FOR{$k = K$ {\bfseries down to} $1$}
            \STATE \textcolor{blue!50}{// Repaint with Alg.~\ref{alg:repaint_algorithm}}
            \STATE $\tilde{\tau}_{k-1} \leftarrow$ \textsc{Repaint}($\tilde{\tau}_k, m, \bar{\alpha}_k, \epsilon_\theta$)
         \ENDFOR
         \STATE $\texttt{Input\_traj} \leftarrow \tilde{\tau}_0$
      \ENDFOR
      
      \STATE $\texttt{Seq\_traj} \leftarrow \tilde{\tau}_0$
      
      \STATE \textcolor{blue!50}{// Post-Processing}
      \STATE Compute actions/rewards for $\texttt{Seq\_traj}$ using $I_\phi, g_\xi$
      \IF{transition validity (Eq.~\eqref{eq:sample_transition}) is satisfied}
         \STATE $D_{\text{aug}} \leftarrow D_{\text{aug}} \cup \{\texttt{Seq\_traj}\}$
      \ENDIF
   \ENDWHILE
   \STATE \textbf{return} $D_{\text{aug}}$
\end{algorithmic}
\end{algorithm}

\begin{algorithm}[H]
   \caption{Repaint}
   \label{alg:repaint_algorithm}
\begin{algorithmic}[1]
   \STATE {\bfseries Input:} Trajectory from previous denoised step $\tilde{\tau}_k$, Mask $m$, Gaussian model $\epsilon_\theta$, pre-defined scheduler $\bar{\alpha}_k$
   
   \STATE {\bfseries Output:} $\tilde{\tau}_{k-1}$
   \STATE $\tilde{\tau}_{k-1}^{\text{prior}} \leftarrow \text{Add noise on } \tilde{\tau}_{k}$ \hfill \COMMENT{Eq.~\eqref{eq:repaint_a}}
   \STATE $\tilde{\tau}_{k-1}^{\text{posterior}} \leftarrow \text{Denoise from }\tilde{\tau}_k$ \hfill \COMMENT{Eq.~\eqref{eq:repaint_b}}
   \STATE $\tilde{\tau}_{k-1} \leftarrow m \odot \tilde{\tau}_{k-1}^{\text{posterior}} + (1-m) \odot \tilde{\tau}_{k-1}^{\text{prior}}$ \hfill \COMMENT{Eq.~\eqref{eq:repaint_c}}
   \STATE \textbf{return} $\tilde{\tau}_{k-1}$
\end{algorithmic}
\end{algorithm}

\newpage{}
\subsection{Loss functions}\label{appendix:training_loss}

\textbf{Diffusion model Training.}~  The model is conditioned on the normalized return $R(\tau)$ to generate observation sequences of length $H$, using the following objective:
\begin{align}
    \mathcal{L}(\theta) = \mathbb{E}_{k,\tau_0\in\mathcal{D},\epsilon,R(\tau),\beta}\Bigl[\bigl\|\epsilon
       - \epsilon_\theta\bigl(\tau_k,k,\,(1-\beta)R(\tau_0) + \beta\emptyset\bigr)\bigr\|^2\Bigr].
\end{align}
where $\tau:=(\boldsymbol{o}_{t:t+H-1})$ is the observation sequence,  $\beta{\sim}\mathrm{Bern}(\lambda)$ is a Bernoulli mask, and $\epsilon{\sim}\mathcal{N}(0,I)$

\textbf{Dynamic models Training.}~ To reconstruct full trajectories including actions and rewards, we train three auxiliary models: an inverse dynamics model $I_\phi(o^i, {o^i}')$ for each agent $i$, a reward model $g_\xi(\boldsymbol{o}, \boldsymbol{a})$, and a transition model $f_\varphi(o^i, a^i)$. These models are optimized jointly to ensure consistency with environment dynamics, using the following objective:
\begin{align}\label{eq:dynamic_module_loss}
    \mathcal{L}(\xi, \phi)  = \mathbb{E}_{\boldsymbol{o},\boldsymbol{a},r,\boldsymbol{o}' \sim D} \Bigl[ \big(r - g_\xi(\boldsymbol{o},\boldsymbol{a})\big)^2 + \sum_{i=1}^{N} \Big(\big(a^i - I_\phi(o^i, {o^i}')\big)^2 + ({o^i}' - f_\varphi(o^i,a^i))^2\Big)\Bigr].
\end{align}

\textbf{Value function Training.}~ 
To mitigate potential estimation bias in the value function, which is critical for determining subgoals, we employ a \emph{\bfseries SARSA}-style objective focused on observed transitions. This approach stabilizes value estimation by reducing extrapolation errors. Furthermore, the generation process remains robust to absolute value bias as it relies on relative ranking for prioritization. Finally, to ensure subgoals reflect the global team return, individual value functions are trained using the collective return, as defined below:
\begin{align}\label{eq:value_loss}
    L(\zeta)=\mathbb{E}_{\boldsymbol{o},r,\boldsymbol{o}' \sim D}\!\Big[\sum_{i=1}^N\big(V_\zeta(o^i) - (r + \gamma V_\zeta({o^i}'))\big)^2\Big].
\end{align}

\newpage
\section{Experimental Details}\label{appendix:experiment_detail}
\subsection{Benchmarks}\label{appendix:mpe_and_smac}

\begin{figure}[ht]
    \centering
    \includegraphics[width=0.9\linewidth]{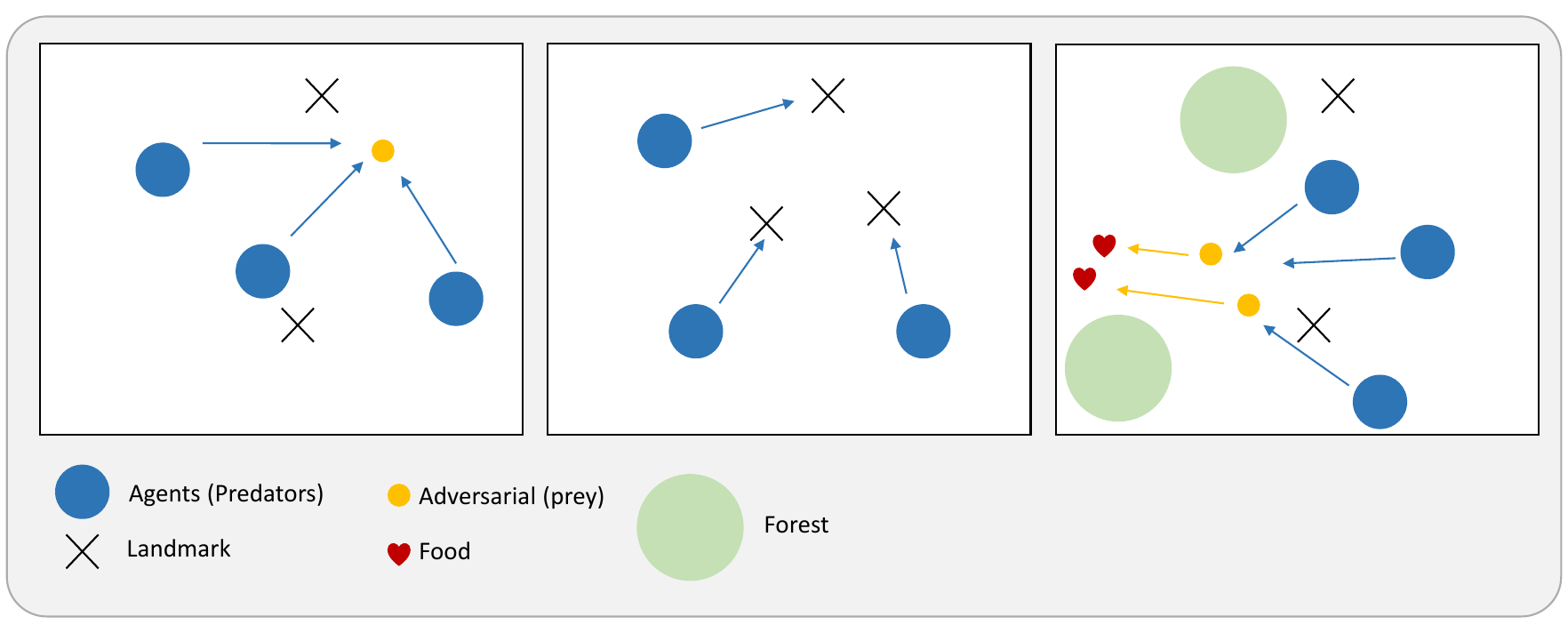}
    \caption{Visualization of MPE tasks. From left to right: Simple Tag, Simple Spread, Simple World.}
    \label{fig:mpe_env}
\end{figure}

\paragraph{MPE.}
\emph{Multi-Agent Particle Environment (MPE~\citep{Lowe17mpe})} is a standard benchmark for continuous multi-agent coordination tasks, featuring simple physical dynamics and inter-agent interactions in a 2D space. We evaluated our method on three representative tasks: \texttt{Tag}, \texttt{Spread}, and \texttt{World}, each requiring distinct coordination.
Fig.~\ref{fig:mpe_env} illustrates the three environments, with detailed descriptions provided below.

\begin{itemize}
    \setlength{\leftskip}{-1.6em}  
    \item \texttt{Simple Tag}: This task involves three slower agents (predators) attempting to capture a faster adversarial agent (prey). Due to their speed disadvantage, predators must coordinate their movements to surround and capture the prey, as individual pursuit is ineffective.

    \item \texttt{Simple Spread}: This environment consists of three agents and three fixed landmarks. Agents are rewarded based on their proximity to the landmarks and penalized for collisions with other agents. The objective is to spread out and cover all landmarks while avoiding collisions, requiring spatial coordination.
    
    \item \texttt{Simple World}: This task involves three cooperative agents and two adversaries. Cooperative agents aim to capture the adversaries, while the adversaries simultaneously try to collect food items scattered throughout the environment and avoid being captured. 
\end{itemize}

\paragraph{SMAC.}
\emph{StarCraft Multi-Agent Challenge (SMAC)} \citep{samvelyan2019starcraft} is a high-dimensional, discrete-action cooperative benchmark built on the StarCraft II engine. In this benchmark, agents must coordinate to defeat enemies using partial observations and task-specific strategies. We evaluated our method on three commonly used SMAC scenarios: \texttt{3m}, \texttt{2s3z}, and \texttt{8m}.
\begin{table}[ht]
\normalsize
\centering
\caption{Agent and enemy compositions for SMAC tasks}
\label{table:env_smac}

\begin{tabular}{l|ccc}
\toprule
\textbf{Task} & \textbf{Agents} & \textbf{vs} & \textbf{Enemy} \\
\midrule
3m & 3 marines & vs & 3 marines \\
2s3z & 2 stalkers and 3 zealots & vs & 2 stalkers and 3 zealots \\
8m & 8 marines & vs & 8 marines \\
\bottomrule
\end{tabular}
\end{table}

\paragraph{SMAC-v2.}
Compared to the original SMAC, which utilizes fixed unit types and deterministic starting configurations, SMAC-v2 \citep{ellis2023smacv2} introduces significantly higher variability. Specifically, both ally and enemy unit compositions are sampled from probability distributions, and initial agent placements are randomized via reflections and spatial perturbations of the map layout. These features increase stochasticity and diversity across episodes, providing a more rigorous testbed for assessing robustness and generalization in multi-agent coordination. A detailed explanation is provided in Appendix~\ref{appendix:smacv2}.

\newpage{}
\subsection{Evaluation Algorithms}\label{appendix:Offline_MARL_algorithms}
This section outlines the evaluation algorithms used for comparison on the original and augmented datasets.
\paragraph{Counterfactual Conservative Q-Learning (CFCQL)~\citep{Shao23cfcql}.}

CFCQL applies conservative regularization to each agent’s Q-function individually in a counterfactual manner. That is, it marginalizes over the actions of other agents while fixing the agent’s own action. This design enables each agent to estimate a conservative Q-value for its own actions while still accounting for the behaviors of others through the environment dynamics. The training objective is formulated as:

\begin{align}
\hat{Q}_{k+1} \leftarrow \arg\min_{Q} \; \alpha \left[ \sum_{i=1}^n \mathbb{E}_{s \sim \mathcal{D},\, a^{-i} \sim \beta^{-i}} \, \mathbb{E}_{a^i \sim \mu^i} \left[ Q(s, a) \right] - \mathbb{E}_{s \sim \mathcal{D},\, a \sim \beta} \left[ Q(s, a) \right] \right] +  \hat{\mathcal{E}}_{\mathcal{D}}(\pi, Q, k).
\label{eq:cfcql}
\end{align}
where $\hat{\mathcal{E}}_{\mathcal{D}}(\pi, Q, k)$ denotes the temporal-difference (TD) error used to fit the Q-function.

\paragraph{Offline Multi-Agent RL with Actor Rectification (OMAR) and MATD3+BC ~\citep{Pan22omar}.}

OMAR introduces a hybrid optimization strategy that combines first-order policy gradients with zeroth-order optimization through population-based search. The actor loss is defined as:
\begin{align}
\mathbb{E}_{\mathcal{D}_i} \left[ (1 - \lambda) Q_i(o_i, \pi_i(o_i)) - \lambda \left\| \pi_i(o_i) - \hat{a}_i \right\|^2 \right].
\label{eq:omar_actor_loss}
\end{align}

where the candidate actions $\hat{a}_i$ are sampled from a population distribution that is adaptively shaped based on the critic $Q_i$. If the candidate actions $\hat{a}_i$ are instead sampled directly from the offline dataset, the algorithm reduces to MATD3+BC.

\subsection{Dataset Composition}
We utilize a combination of established offline MARL repositories and custom-generated datasets to evaluate MASTARS across varying levels of environmental complexity.

\paragraph{MPE and SMAC.}
For the Multi-Agent Particle Environment (MPE) and the original StarCraft Multi-Agent Challenge (SMAC), we adopted the standardized datasets provided by off-the-grid MARL (og-marl) \citep{Formanek23ogmarl}. To effectively investigate the effects of the augmented data, we used a smaller dataset compared to the original offline dataset. The dataset size for each task is as follows: each MPE task contains 2,000 episodes with a trajectory length of 25, and each SMAC task contains approximately 250,000 transition samples $(s, a, r, s')$.
\paragraph{SMAC-v2.}
Unlike the original SMAC, for which standardized datasets are readily available, standardized offline datasets for SMAC-v2 are not yet widely available in existing repositories like og-marl. To address this, we generated custom trajectory datasets for various tasks, each consisting of 2,000 episodes with a fixed horizon of 200 steps. These trajectories were collected using the QMIX algorithm \citep{qmix} implemented within the PyMARL framework. For a comprehensive description of the generation process and hyperparameters, please refer to Appendix~\ref{appendix:smacv2}.

\newpage{}
\section{SMAC-v2}
\label{appendix:smacv2}
This appendix provides additional details on the environment, the composition of the dataset, and extended experimental results on the more challenging SMAC-v2 \citep{ellis2023smacv2} benchmark to further evaluate the generalizability of MASTARS beyond the settings considered in the main experiments.

Compared to SMAC, which uses fixed unit types and deterministic starting configurations, SMAC-v2 introduces substantially greater variability: both ally and enemy unit compositions are sampled from probability distributions, and initial agent placements are randomized through reflections and spatial perturbations of the map layout. These features make SMAC-v2 more stochastic and diverse across episodes, providing a stronger testbed for assessing robustness and generalization in multi-agent coordination.

Table~\ref{tab:smacv2_unit_dist} summarizes the unit-type sampling distributions used to generate our datasets for each faction (Protoss, Terran, and Zerg). These stochastic unit compositions, combined with randomized starting positions, provide diverse combat scenarios that challenge the robustness and generalization ability of multi-agent reinforcement learning methods.
\begin{table}[ht]
\centering
\small
\caption{Unit-type sampling probabilities and starting-position distributions for SMAC-v2 datasets.}
\label{tab:smacv2_unit_dist}
\begin{tabular}{l|c|c|c}
\toprule
\textbf{Faction} & \textbf{Unit Type} & \textbf{Probability} & \textbf{Start Position (p=0.5)} \\
\midrule
\multirow{3}{*}{Protoss} 
    & Stalker  & 0.45 & \multirow{3}{*}{Surrounded / Reflected} \\
    & Zealot   & 0.45 &  \\
    & Colossus & 0.10 &  \\
\midrule
\multirow{3}{*}{Terran} 
    & Marine    & 0.45 & \multirow{3}{*}{Surrounded / Reflected} \\
    & Marauder  & 0.45 &  \\
    & Medivac   & 0.10 &  \\
\midrule
\multirow{3}{*}{Zerg} 
    & Zergling  & 0.45 & \multirow{3}{*}{Surrounded / Reflected} \\
    & Baneling  & 0.45 &  \\
    & Hydralisk & 0.10 &  \\
\bottomrule
\end{tabular}
\end{table}

To construct the SMAC-v2 datasets, we collected 2,000 episodes for each task, with all episodes having a fixed horizon of 200 steps. To obtain these trajectories, we used the QMIX algorithm \citep{qmix}, a widely adopted baseline in multi-agent reinforcement learning, implemented in the PyMARL framework, which is a widely adopted baseline in multi-agent reinforcement learning.

Due to the increased stochasticity and difficulty of SMAC-v2, the collected datasets do not achieve near-expert performance (i.e., a test win rate close to 1.0). Instead, we provide two dataset qualities, medium and good, reflecting different levels of policy competence under these challenging settings. Detailed statistics for each task, including the number of trajectories, average return, and test win rate, are reported in Table~\ref{tab:smacv2_dataset}.

\begin{table}[ht]
    \centering
    
    \caption{Dataset statistics for SMAC-v2 tasks.}
    \label{tab:smacv2_dataset}
        \small
    \setlength{\tabcolsep}{6pt}      
    \renewcommand{\arraystretch}{0.9} 
    \begin{tabular}{l l r r r}
        \toprule
        \textbf{Task} & \textbf{Quality} & \textbf{\# Trajectories} & \textbf{Average return} & \textbf{Average win rate} \\
        \midrule
        \multirow{2}{*}{protoss 3\_vs\_3}
        & medium          & 2000  & 16.44 & 0.42 \\
        & good          & 2000  & 18.16 & 0.58 \\
        \midrule
        \multirow{2}{*}{protoss 5\_vs\_5}
        & medium          & 2000  & 17.98  & 0.41 \\
        & good          & 2000  & 19.38 & 0.64 \\
        \midrule
        \multirow{2}{*}{terran 3\_vs\_3}
        & medium          & 2000  & 11.6 & 0.44 \\
        & good          & 2000  & 16.2 & 0.63 \\
        \midrule
        \multirow{2}{*}{terran 5\_vs\_5}
        & medium          & 2000  & 13.09 & 0.4 \\
        & good          & 2000  & 12.26 & 0.8 \\
        \midrule
        \multirow{2}{*}{zerg 3\_vs\_3}
        & medium          & 2000  & 9.75  & 0.43 \\
        & good          & 2000  & 13.33 & 0.57 \\
        \midrule
        \bottomrule
    \end{tabular}
\end{table}

Table~\ref{tab:smacv2_performance} summarizes the performance of all methods on the SMAC-v2 tasks. MASTARS consistently outperforms existing baselines across both CFCQL and OMAR. Notably, the synthetic data generated by MASTARS leads to substantial performance improvements, achieving a 143.7\% increase under CFCQL and a 189.1\% increase under OMAR. Furthermore, MASTARS ranks within the top-two methods on 8 out of 10 tasks under CFCQL, and achieves a top-two ranking on all tasks under OMAR, reflecting its robustness across diverse scenarios and evaluation settings.

These results suggest that MASTARS effectively captures the underlying distribution of complex multi-agent interactions, even in highly stochastic environments where standard offline methods often struggle with distributional shift. The consistent performance gain across different base algorithms (CFCQL and OMAR) further demonstrates that our framework provides a universally beneficial data augmentation strategy that is not tied to a specific policy optimization objective.

\begin{table}[ht]
    \normalsize
    \centering
    \caption{Performance on SMAC-v2 tasks (\textbf{bold}: best, \underline{underline}: second best). Mean $\pm$ std.}
    \label{tab:smacv2_performance}
    \begin{adjustbox}{width=\linewidth,center}
    \robustify\bfseries
    \renewcommand{\arraystretch}{1.5}
    \begin{tabular}{@{}l||l|cc|cc|cc|cc|cc||c@{}}
    \toprule
    \multicolumn{2}{@{}l@{}}{} &
    \multicolumn{2}{c}{Protoss 3\_vs\_3} &
    \multicolumn{2}{c}{Protoss 5\_vs\_5} &
    \multicolumn{2}{c}{Terran 3\_vs\_3} &
    \multicolumn{2}{c}{Terran 5\_vs\_5} &
    \multicolumn{2}{c}{Zerg 3\_vs\_3} &
    \multicolumn{1}{c@{}}{} \\

    \cmidrule(lr){3-4}\cmidrule(lr){5-6}\cmidrule(lr){7-8}\cmidrule(lr){9-10}\cmidrule(lr){11-12}

    \textbf{Alg} & \textbf{Dataset} & Medium & Good & Medium & Good & Medium & Good & Medium & Good & Medium & Good & Average \\

    \midrule
    \multirow{6}{*}{CFCQL}
    & Original 
        & 0$\pm$0 & 0.31$\pm$0.05 & 0.14$\pm$0.07 & 0.18$\pm$0.07 & 0.23$\pm$0.16 & 0.40$\pm$0.55 & 0.15$\pm$0.25 & 0.15$\pm$0.21 & 0.20$\pm$0.45 & \underline{0.30$\pm$0.17} & 0.21 \\
    & MADiff* 
        & 0.26$\pm$0.16 & 0.27$\pm$0.15 & 0.06$\pm$0.09 & 0.10$\pm$0.24 & 0.39$\pm$0.34 & \underline{0.56$\pm$0.44} & 0.23$\pm$0.04 & 0.14$\pm$0.19 & 0.10$\pm$0.10 & 0.25$\pm$0.18 & 0.24 \\
    & DoF* 
        & \textbf{0.59$\pm$0.38} & \underline{0.46$\pm$0.51} & 0.29$\pm$0.42 & 0.14$\pm$0.09 & \textbf{0.52$\pm$0.46} & 0.39$\pm$0.15 & 0.18$\pm$0.17 & 0.32$\pm$0.41 & 0.25$\pm$0.43 & 0.25$\pm$0.43 & \underline{0.34} \\
    & MADiTS
        & 0.34$\pm$0.39 & 0.33$\pm$0.41 & 0.09$\pm$0.07 & 0.05$\pm$0.06 & 0.20$\pm$0.45 & 0.36$\pm$0.07 & \underline{0.28$\pm$0.41} & 0.02$\pm$0.04 & \textbf{0.26$\pm$0.43} & 0.28$\pm$0.12 & 0.22 \\
    & INS
        & 0.16$\pm$0.13 & 0.40$\pm$0.52 & \underline{0.42$\pm$0.52} & \underline{0.20$\pm$0.45} & \underline{0.44$\pm$0.25} & 0.45$\pm$0.36 & \textbf{0.40$\pm$0.55} & \underline{0.45$\pm$0.51} & 0.23$\pm$0.09 & 0.20$\pm$0.45 & \underline{0.34} \\
    & Ours
        & \underline{0.50$\pm$0.52} & \textbf{0.48$\pm$0.71} & \textbf{0.50$\pm$0.71} & \textbf{0.28$\pm$0.45} & 0.21$\pm$0.07 & \textbf{0.91$\pm$0.00} & 0.18$\pm$0.09 & \textbf{1.00$\pm$0.00} & \textbf{0.26$\pm$0.04} & \textbf{0.70$\pm$0.50} & \textbf{0.50} \\

    \midrule
    \multirow{6}{*}{OMAR}
    & Original
        & 0$\pm$0 & 0.13$\pm$0.07 & 0.10$\pm$0.06 & 0.06$\pm$0.04 & 0.12$\pm$0.07 & 0.28$\pm$0.22 & \underline{0.31$\pm$0.40} & 0.13$\pm$0.12 & \underline{0.22$\pm$0.43} & \textbf{0.21$\pm$0.19} & 0.16 \\
    & MADiff*
        & 0.15$\pm$0.08 & 0.16$\pm$0.13 & \underline{0.21$\pm$0.44} & 0.01$\pm$0.01 & \underline{0.48$\pm$0.48} & \textbf{0.37$\pm$0.36} & 0.11$\pm$0.07 & 0.08$\pm$0.08 & 0.06$\pm$0.06 & 0.12$\pm$0.09 & \underline{0.18} \\
    & DoF*
        & 0.12$\pm$0.1 & 0.06$\pm$0.1 & 0.08$\pm$0.09 & 0.06$\pm$0.09 
        & 0.05$\pm$0.1  &0.32$\pm$0.39 & 0.30$\pm$0.40 & 0.24$\pm$0.43 & 0.20$\pm$0.45 & 0.03$\pm$0.05 & 0.15 \\
    & MADiTS
        & \underline{0.25$\pm$0.43} & 0.06$\pm$0.07 & 0.12$\pm$0.07 & \underline{0.21$\pm$0.44} & 0.20$\pm$0.45 & 0.10$\pm$0.08 & 0.20$\pm$0.45 & 0.03$\pm$0.05 & 0.13$\pm$0.03 & 0.07$\pm$0.03 & 0.14 \\
    & INS
        & 0.08$\pm$0.07 & \underline{0.20$\pm$0.45} & 0.20$\pm$0.45 & 0$\pm$0 & 0.21$\pm$0.04 & 0.18$\pm$0.05 & 0$\pm$0 & \textbf{0.45$\pm$0.48} & 0.14$\pm$0.23 & 0$\pm$0 & 0.15 \\
    & Ours
        & \textbf{0.55$\pm$0.64} & \textbf{0.65$\pm$0.51} & \textbf{0.39$\pm$0.53} & \textbf{0.61$\pm$0.45} & \textbf{0.52$\pm$0.41} & \underline{0.31$\pm$0.55} & \textbf{0.55$\pm$0.36} & \underline{0.29$\pm$0.48} & \textbf{0.46$\pm$0.06} & \underline{0.18$\pm$0.12} & \textbf{0.45} \\
    \bottomrule
    \end{tabular}
    \end{adjustbox}
\end{table}

\newpage
\section{Implementation Details and Hyperparameters}\label{appendix:implementation_detail}
We present the implementation details and hyperparameter settings that affect the performance and stability of our models during offline training.

\paragraph{Architectural and Experimental Setup.}
For the diffusion model, we adopt the same hyperparameter configuration as MADiff~\citep{Zhu024madiff}. All auxiliary networks, including the value function, inverse dynamics model, reward model, and transition model, are implemented as two-layer multilayer perceptrons (MLPs) with ReLU activations and 256 hidden units per layer. In addition, the models, such as the diffusion model, value function, and dynamic models, were trained over 100,000 time steps. All experiments were conducted with five random seeds to ensure statistical reliability with the main results reported as mean $\pm$ standard deviation.

\paragraph{Fairness of Comparison.}
All augmentation methods are compared under matched budgets. Each method generates 1,000 episodes; INS is matched by the total number of transitions, since it synthesizes transitions rather than trajectories. MADiff*, DoF*, MADiTS, and MASTARS share the same post-processing filter (Eq.~\eqref{eq:sample_transition}), the same auxiliary models, and the same number of diffusion steps, whereas INS retains its native synthesis and filtering protocol. All methods use the same training and tuning budget, including the same search space for the acceptance threshold $\epsilon$, and we report the best validation result of each method. Because the amount of accepted augmented data is fixed, a looser threshold does not provide more training data to a baseline. All other settings follow the recommended hyperparameters in the original papers.

\paragraph{Hyperparameters on Sequential Inpainting.}
 Across all experiments, including MPE, SMAC, and SMAC-v2, we generate 1,000 augmented trajectories, each consisting of 8 timesteps (i.e., $H=8$). The diffusion model performs 200 denoising steps during sampling for both MADiff* and MASTARS. For MADiTS, we follow the original hyperparameter settings from the paper for training and sampling, except for the adaptive threshold and the number of generated trajectories, which are adjusted to ensure a fair comparison.

\newpage
\paragraph{Hyperparameters for the Post-Processing.}
This section describes the threshold values used in the post-processing module to filter out invalid transitions based on Eq.~\eqref{eq:sample_transition}. 

The main results on the MPE environment (Table~\ref{table:mpe_main_result}) are obtained using task-specific adaptive thresholds for all baselines, including our method, as specified in Table~\ref{table:mpe_threshold}. Since INS does not employ thresholds for trajectory generation, we exclude it from this setting. 

For the ablation studies, we use the same thresholds as those reported in Table~\ref{table:mpe_threshold}, except for the w/o subgoal variant. When generating trajectories without subgoals, applying low $\epsilon$ values for certain tasks results in overly strict acceptance criteria, leading to no trajectories being generated. To address this issue, we adopt slightly higher $\epsilon$ values for Tag Random, World Medium, World MD-Replay, and World Random, setting them to 0.03, 0.004, 0.02, and 0.02, respectively.

For the SMAC and SMAC-v2 benchmarks, MADiff*, DoF*, and MADiTS, use higher thresholds than MASTARS, as detailed in Tables~\ref{table:smac_threshold} and \ref{table:smacv2_threshold}, since they generate entire trajectories, resulting in samples that deviate more significantly from the original dataset distribution. In contrast, our method only generates trajectory segments after the subgoal timestep, enabling the use of lower thresholds and stricter acceptance criteria.

\begin{table}[ht]
\centering
\small
\caption{MPE thresholds in all algorithms}
\label{table:mpe_threshold}
\begin{tabular}{l|cccc}
\toprule
\textbf{Task \textbackslash{} Algo} & \textbf{MADiff*} & \textbf{DoF*} & \textbf{MADiTS} & \textbf{Ours} \\
\midrule
Tag-Medium     & 0.03   & 0.02  & 0.03 & 0.008 \\
Tag-MD-Replay  & 0.03   & 0.02  & 0.03 & 0.006 \\
Tag-Random     & 0.04   & 0.03  & 0.03 & 0.007 \\
Spread-Medium     & 0.02   & 0.02  & 0.03 & 0.01 \\
Spread-MD-Replay  & 0.02   & 0.02  & 0.03 & 0.004\\
Spread-Random     & 0.03   & 0.02  & 0.03 & 0.001\\
World-Medium     & 0.04   & 0.04  & 0.01 & 0.001 \\
World-MD-Replay  & 0.02   & 0.04  & 0.03 & 0.003\\
World-Random     & 0.04   & 0.04  & 0.03 &0.009\\
\bottomrule
\end{tabular}
\end{table}

\begin{table}[ht]
\centering
\small
\caption{SMAC thresholds in all algorithms}
\label{table:smac_threshold}
\begin{tabular}{l|cccc}
\toprule
\textbf{Task \textbackslash{} Algo} & \textbf{MADiff*} & \textbf{DoF*} & \textbf{MADiTS} & \textbf{Ours}\\
\midrule
3m-Medium     & 0.3  & 0.05  & 1.0  & 0.3 \\
3m-Good       & 0.3  & 0.1   & 1.0  & 0.002 \\
2s3z-Medium   & 0.7  & 0.3   & 2.0  & 0.09 \\
2s3z-Good     & 0.7  & 0.4   & 2.0  & 0.009 \\
8m-Medium     & 1.0  & 0.4   & 2.0  & 0.2 \\
8m-Good       & 1.0  & 0.8   & 2.0  & 0.3 \\
\bottomrule
\end{tabular}
\end{table}

\begin{table}[ht]
\centering
\small
\caption{SMAC-v2 thresholds in all algorithms}
\label{table:smacv2_threshold}
\begin{tabular}{l|cccc}
\toprule
\textbf{Task \textbackslash{} Algo} & \textbf{MADiff*} & \textbf{DoF*} & \textbf{MADiTS} & \textbf{Ours} \\
\midrule
Protoss 3\_vs\_3-Medium  & 1.0  & 0.5  & 1.0  & 1.0 \\
Protoss 3\_vs\_3-Good    & 1.0  & 0.8  & 1.2  & 0.02 \\
Protoss 5\_vs\_5-Medium  & 1.0  & 0.5  & 1.0  & 0.9 \\
Protoss 5\_vs\_5-Good    & 1.0  & 0.8  & 1.2  & 0.9 \\
Terran 3\_vs\_3-Medium   & 1.0  & 0.5 & 0.9  & 0.02 \\
Terran 3\_vs\_3-Good     & 1.0  & 0.7 & 1.2  & 1.0 \\
Terran 5\_vs\_5-Medium   & 1.0  & 0.5 & 0.9  & 0.6 \\
Terran 5\_vs\_5-Good     & 1.0  & 0.8 & 1.2  & 0.1 \\
Zerg 3\_vs\_3-Medium     & 1.0  & 0.5 & 0.9  & 0.7 \\
Zerg 3\_vs\_3-Good       & 1.0  & 0.6 & 0.8  & 0.01 \\
\bottomrule
\end{tabular}
\end{table}

\clearpage
\section{Computational Cost}\label{appendix:computational_cost}
All experiments, including model training and data generation, were conducted on a single machine with 8 NVIDIA GeForce RTX 3090 GPUs, 96 logical CPU cores, and Linux (kernel 5.15), using Python 3.8 with CUDA acceleration. Only the wall-clock timings reported in Tables~\ref{tab:training_cost}--\ref{tab:partial_cost} were measured separately on a single RTX 4090 with batch size 256 and $K=200$ denoising steps.

\subsection{Training Cost}\label{appendix:training_cost}
To evaluate computational efficiency, we compared the training time of MASTARS against the MADiTS baseline on the MPE Simple Spread (MD-Replay) task on a single RTX 4090. Both methods were trained for 100K steps. MADiTS required approximately 3 hours and 7 minutes (187 minutes), while MASTARS completed training in 2 hours and 20 minutes (140 minutes), achieving a ${\sim}25\%$ reduction in training time under identical conditions. Since MASTARS and MADiff have the same training architecture, MADiff incurs the same training cost as MASTARS.

The four auxiliary models (reward, inverse dynamics, transition, and value) are lightweight two-layer MLPs (Appendix~\ref{appendix:implementation_detail}) and are trained jointly with the diffusion model in the same training loop rather than in separate runs. As summarized in Table~\ref{tab:training_cost}, they add only 0.3M parameters, about 0.5\% of the 64.3M-parameter diffusion model, so their training overhead is negligible.

\begin{table}[ht]
    \centering
    \caption{Training cost of MASTARS on a single RTX 4090}
    \label{tab:training_cost}
    \renewcommand{\arraystretch}{1.2}
    \begin{adjustbox}{width=0.75\linewidth,center}
    \robustify\bfseries
    \begin{tabular}{l|cc}
        \toprule
        Training block & Parameters & Measured training \\
        \midrule
        Diffusion model       & 64.3M        & 100K steps: 2.33 h (0.084 s/step) \\
        Four auxiliary models & 0.3M (0.5\%) & Same training loop \\
        \bottomrule
    \end{tabular}
    \end{adjustbox}
\end{table}

\subsection{Generation Cost}\label{appendix:generation_cost}
MASTARS performs $N$ agent-wise inpainting stages of $K$ denoising steps, so its per-candidate sampling cost grows with the number of agents. Since rejected candidates never enter the dataset, however, the practical cost is the \emph{time per accepted trajectory} (candidate latency divided by the acceptance rate). Table~\ref{tab:accepted_cost} reports it for joint, simple sequential, and MASTARS generation (all built on MADiff; Appendix~\ref{appendix:madiff_data_generation}) under three acceptance thresholds $\epsilon$.

\begin{table}[ht]
    \centering
    \caption{Time per accepted trajectory in MPE under different acceptance thresholds $\epsilon$ on a single RTX 4090. $\infty$: no candidate accepted. \textbf{Bold}: fastest per setting.}
    \label{tab:accepted_cost}
    \renewcommand{\arraystretch}{1.2}
    \begin{adjustbox}{width=0.6\linewidth,center}
    \robustify\bfseries
    \begin{tabular}{l|l|ccc}
        \toprule
        Env & Method & $\epsilon=0.001$ & $\epsilon=0.005$ & $\epsilon=0.01$ \\
        \midrule
        \multirow{3}{*}{Spread} & Joint      & 0.50s & \textbf{0.05s} & \textbf{0.04s} \\
                                & Simple-seq & 1.52s & 0.13s & 0.12s \\
                                & MASTARS    & \textbf{0.26s} & 0.13s & 0.12s \\
        \midrule
        \multirow{3}{*}{Tag}    & Joint      & $\infty$ & 0.55s & \textbf{0.12s} \\
                                & Simple-seq & $\infty$ & 1.36s & 0.32s \\
                                & MASTARS    & \textbf{0.51s} & \textbf{0.26s} & 0.19s \\
        \midrule
        \multirow{3}{*}{World}  & Joint      & $\infty$ & 0.43s & \textbf{0.15s} \\
                                & Simple-seq & $\infty$ & 1.18s & 0.48s \\
                                & MASTARS    & \textbf{0.53s} & \textbf{0.29s} & 0.21s \\
        \bottomrule
    \end{tabular}
    \end{adjustbox}
\end{table}

Joint generation is fastest at the loose threshold $\epsilon=0.01$, but its acceptance rate collapses as $\epsilon$ tightens: at $\epsilon=0.001$ it accepts no candidate on Tag and World. MASTARS matches or outperforms simple sequential generation at every threshold owing to its higher acceptance rate, and is thus the most efficient option whenever validation is strict.

\subsection{Scaling with the Number of Agents and Horizon}\label{appendix:generation_scaling}
Table~\ref{tab:latency_scaling} reports the per-candidate sampling latency in SMAC as the number of agents grows (3m, 5m\_vs\_6m, 8m, and 25m with 3, 5, 8, and 25 agents) and as the horizon $H$ grows on 3m. Latency varies only modestly with the horizon, since one diffusion pass covers all timesteps, but increases sharply with the number of agents, which require sequential inpainting passes. This reveals a computation--performance trade-off: MASTARS incurs a higher generation cost as the agent count grows, but it also delivers the strongest downstream performance in the largest-agent and longest-horizon settings (Table~\ref{tab:scalability}).

\begin{table}[ht]
    \centering
    \caption{Per-candidate sampling latency of MASTARS in SMAC on a single RTX 4090}
    \label{tab:latency_scaling}
    \renewcommand{\arraystretch}{1.2}
    \small
    \begin{minipage}[t]{0.5\linewidth}
        \centering
        \begin{tabular}{l|cccc}
            \toprule
            Map      & 3m   & 5m\_vs\_6m & 8m   & 25m \\
            \midrule
            \#agents & 3    & 5          & 8    & 25  \\
            Time (s) & 0.12 & 0.30       & 0.75 & 8.3 \\
            \bottomrule
        \end{tabular}\\[0.9em]
        (a) Number of agents
    \end{minipage}%
    \begin{minipage}[t]{0.5\linewidth}
        \centering
        \begin{tabular}{l|cccc}
            \toprule
            Map      & 3m   & 3m   & 3m   & 3m \\
            \midrule
            Horizon $H$ & 8    & 16   & 24   & 32 \\
            Time (s)    & 0.29 & 0.25 & 0.29 & 0.31 \\
            \bottomrule
        \end{tabular}\\[0.9em]
        (b) Horizon length
    \end{minipage}
\end{table}

\subsection{Cost Reduction from Partial Generation}\label{appendix:partial_cost}
Partial generation preserves each trajectory up to its value-based subgoal and synthesizes only the suffix; on average only 44\% of a trajectory is synthesized (Appendix~\ref{appendix:synthesis_fraction}). To verify that this reduced search space translates into lower cost, we measure the end-to-end cost of obtaining 1,000 trajectories accepted by Eq.~\eqref{eq:sample_transition} with and without subgoals, using the same source data, threshold, hardware, and seeds, averaged over the three MPE environments. Both variants take 0.12s per candidate, so the difference comes entirely from the acceptance rate. As shown in Table~\ref{tab:partial_cost}, partial generation reduces the number of failed attempts and thereby the wall-clock time by 40\%.

\begin{table}[ht]
    \centering
    \caption{Cost of collecting 1,000 accepted trajectories with and without subgoal-based partial generation on a single RTX 4090}
    \label{tab:partial_cost}
    \renewcommand{\arraystretch}{1.2}
    \setlength{\tabcolsep}{12pt}
    \begin{adjustbox}{width=\linewidth,center}
    \robustify\bfseries
    \begin{tabular}{l|cccc}
        \toprule
        Variant & \makecell{Synthesized\\fraction} & \makecell{Attempts for\\1,000 accepted} & \makecell{Rejection\\rate} & \makecell{Wall-clock\\time} \\
        \midrule
        w/o Subgoal & 1.00 & 2,732 & 63\% & 328s \\
        MASTARS     & 0.44 & 1,623 & 38\% & \textbf{195s} \\
        \bottomrule
    \end{tabular}
    \end{adjustbox}
\end{table}

\newpage{}
\section{Additional Experimental Analyses}
\label{appendix:additional_experiments}

\subsection{Effect of Hyperparameters} 
\label{appendix:hyperparameter}

\begin{figure}[htbp]
    \centering
    \begin{subfigure}[t]{0.48\textwidth}
        \centering
        \includegraphics[width=\linewidth]{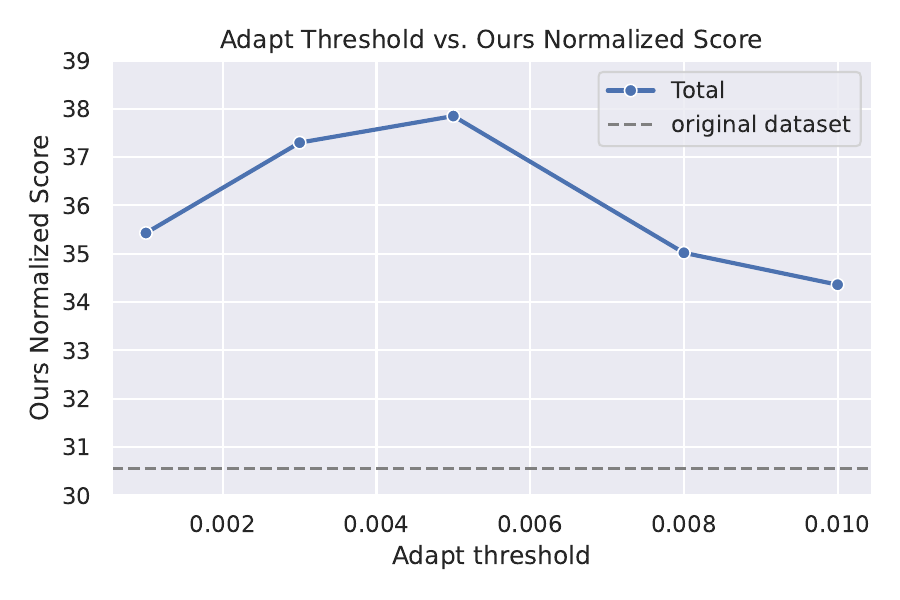}
        \caption{Performance across acceptance threshold $\epsilon$.}
        \label{fig:hyperparam_sensitivity}
    \end{subfigure}
    \hfill
    \begin{subfigure}[t]{0.48\textwidth}
        \centering
        \includegraphics[width=\linewidth]{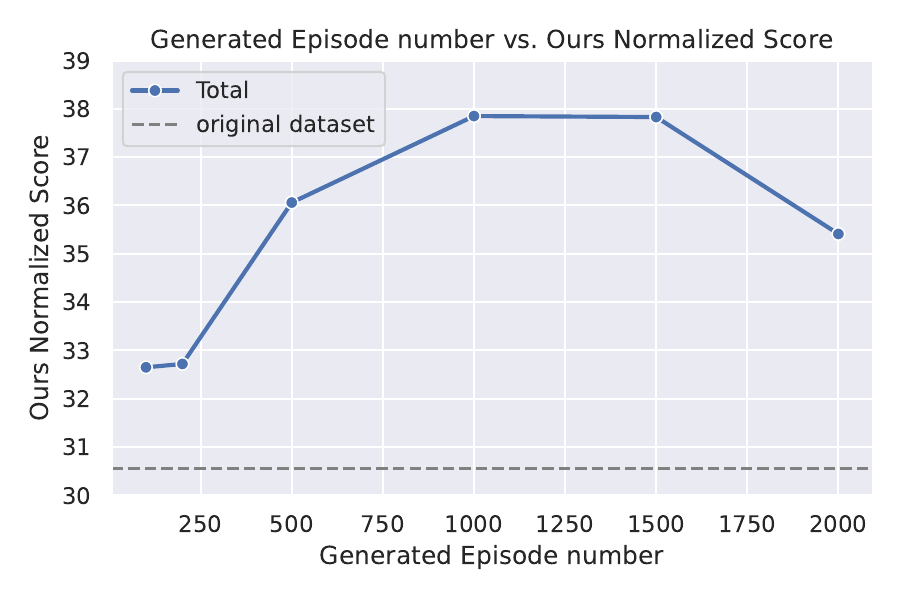}
        \caption{Performance with different amounts of generated trajectories.}
        \label{fig:num_generated}
    \end{subfigure}
    \caption{Analysis of generation sensitivity. (a) shows that our method is robust to threshold selection. (b) shows that moderate augmentation leads to better performance.}
    \label{fig:hyperparam_and_quantity}
\end{figure}

We analyze the sensitivity of MASTARS to two key hyperparameters: the acceptance threshold $\epsilon$ and the number of generated trajectories, both of which influence the diversity of augmented data. For the analysis of the number of generated trajectories, we fix $\epsilon$ to 0.005. As shown in Fig.~\ref{fig:hyperparam_and_quantity}, MASTARS is robust to variations in $\epsilon$, while the number of generated trajectories has a more significant impact. Generating too few samples yields limited improvement, whereas excessive augmentation can introduce distributional shift and degrade performance. Optimal results are achieved with a moderate amount of augmentation, highlighting the importance of balancing quantity and quality. Notably, MASTARS consistently outperforms the original dataset across all settings.

\vspace{2em}

\subsection{Synthesis Fraction and Efficiency}\label{appendix:synthesis_fraction}
To verify that MASTARS avoids full-sequence regeneration, we measured the proportion of trajectories synthesized via value-based subgoals. By bypassing the excessively large search space of full regeneration and focusing exclusively on necessary refinements, MASTARS synthesizes only 44\% of each trajectory on average, as reported in  Table~\ref{tab:percent_subgoal}. This demonstrates an effective balance between leveraging high-value existing segments and generating new segments to ensure team-wide coordination.

\begin{table}[ht]
    \caption{Post-subgoal ratio}
    \label{tab:percent_subgoal}
    \renewcommand{\arraystretch}{1.1}
    \begin{adjustbox}{width=0.5\linewidth,center}
    \robustify\bfseries
    \begin{tabular}{l|ccc}
    \toprule
    ~ & Medium & MD-Replay & Random \\
    \midrule
    Tag    & 0.42 & 0.52 & 0.63 \\
    Spread & 0.58 & 0.26 & 0.62 \\
    World  & 0.44 & 0.17 & 0.35 \\
    \bottomrule
    \end{tabular}
    \end{adjustbox}
\end{table}

\newpage
\subsection{Scalability Analysis w.r.t. Agent Number and Horizon Length}  
\label{appendix:scalability}
To verify scalability in terms of agent number and horizon length, we conducted additional experiments in SMAC: the 25m scenario (25 agents) for agent scalability, and the 2c\_vs\_64zg scenario (horizon length 400) for temporal scalability. For both settings, we report results averaged over CFCQL and OMAR. As shown in Table~\ref{tab:scalability}, MADiTS showed performance drops in these challenging scenarios, while MASTARS consistently outperformed the original offline datasets. These results highlight the effective scalability of MASTARS, demonstrating strong performance even in scenarios with many agents and extended horizons.
\vspace{0.5em}
\begin{table}[ht]
    \centering
    \caption{Test win rate on 25m and 2c\_vs\_64zg in the SMAC environment}
    \label{tab:scalability}
    \renewcommand{\arraystretch}{1.4} 
    \begin{adjustbox}{width=.8\linewidth,center}
	\robustify\bfseries
    \begin{tabular}{l|cccc}
    \hline
    \textbf{Task \textbackslash Algorithms } & Original & MADiff* & MADiTS & MASTARS (Ours) \\
    \hline
    25m & 0.7 & 0.75 & 0.6 & \textbf{0.87} \\
    2c\_vs\_64zg & 0.52 & 0.7 & 0.05 & \textbf{0.87} \\
    \hline
    \end{tabular}
    \end{adjustbox}
\end{table}

\subsection{Sensitivity Analysis of Value Estimation}\label{appendix:value_sensitivity}
\begin{figure}[ht]
    \centering
    \includegraphics[width=0.6\linewidth]{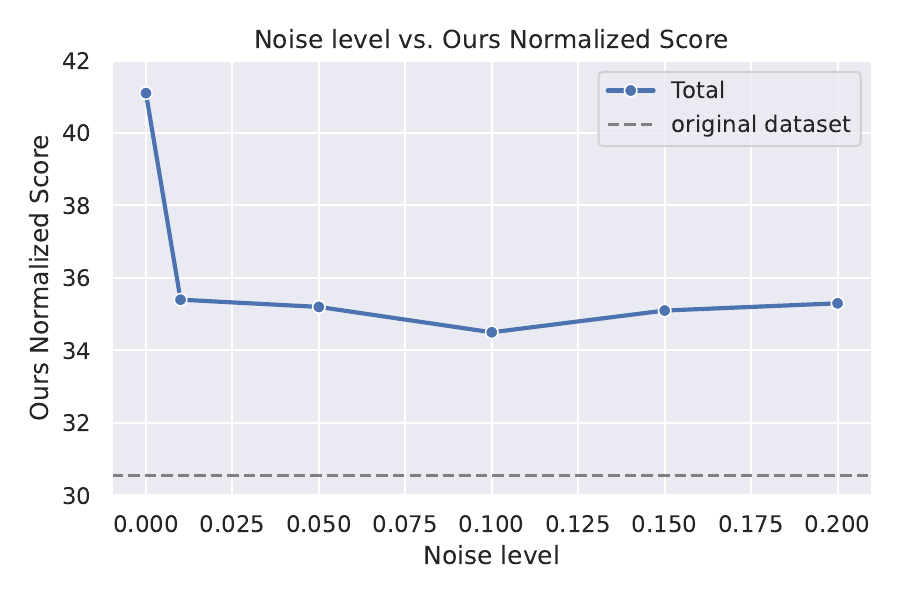}
    \caption{Average performance across different value noise levels}
    \label{fig:noise_value}
\end{figure}

We tested the stability of our SARSA-style value learning. Across MPE benchmarks, normalized TD error steadily converged to near zero (within $\sim$1\%), as value learning progressed, confirming reliable estimation (Table~\ref{tab:value_loss}). To test sensitivity, we intentionally injected random noise into values during subgoal generation, perturbing each value within $[(1-\sigma)\text{value}, (1+\sigma)\text{value}]$. As Fig.~\ref{fig:noise_value} shows, performance dropped with noise, underscoring the need for order-preserving estimates. Still, for all tested noise levels up to $\sigma=0.2$, the score remained within 34.50--35.40 and MASTARS consistently outperformed the original dataset (30.56).
\begin{table}[ht]
    \centering
    \caption{TD error for value estimation}
    \label{tab:value_loss}
    \renewcommand{\arraystretch}{1.4} 
    \begin{adjustbox}{width=0.4\linewidth,center}
    \robustify\bfseries
    \begin{tabular}{l|cccc}
    \hline
    \textbf{Time} & 0 & 30K & 60K & 90K \\
    \hline
    Value Loss & 86.3 & 0.02 & 0.01 & 0.01 \\
    \hline
    \end{tabular}
    \end{adjustbox}
\end{table}

\newpage
\subsection{Robustness to Reward and Inverse-Dynamics Errors}\label{appendix:annotation_noise}
\begin{figure}[ht]
    \centering
    \includegraphics[width=0.6\linewidth]{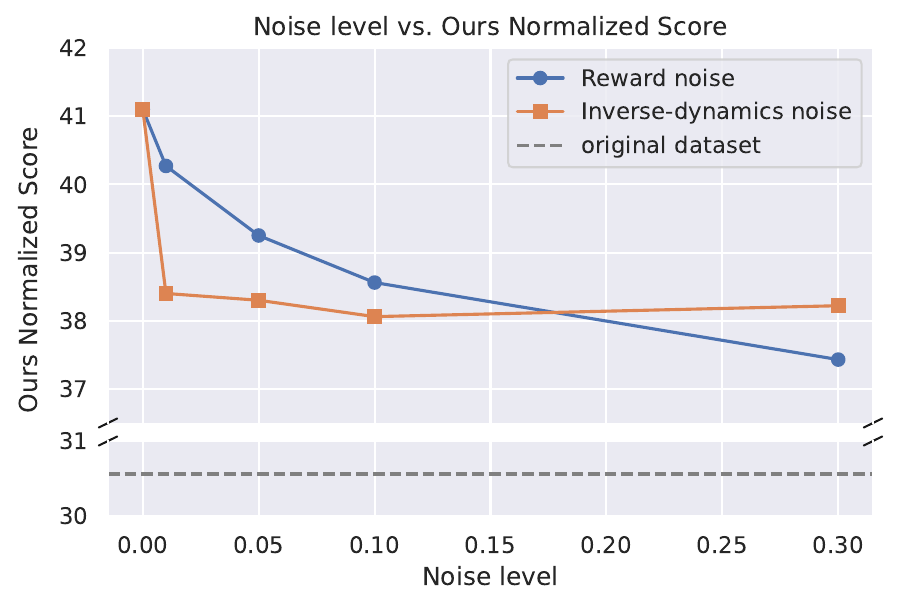}
    \caption{Average performance across different reward and inverse-dynamics noise levels}
    \label{fig:annotation_noise}
\end{figure}

The reward model $g_\xi$ and inverse dynamics model $I_\phi$ only annotate generated observation sequences. Following the value-noise protocol in Appendix~\ref{appendix:value_sensitivity}, we perturb every predicted reward or action within $[(1-\sigma)x, (1+\sigma)x]$ before adding it to the augmented dataset. Scores are averaged over all MPE environments, dataset levels, and offline MARL algorithms; $\sigma=0$ denotes unmodified MASTARS and Original (30.56) denotes training on the unaugmented dataset. As shown in Fig.~\ref{fig:annotation_noise}, reward noise gradually lowers the score to 37.43, while inverse-dynamics noise causes an initial drop but stays at 38.06--38.40. All noisy settings outperform Original, showing graceful degradation.

\clearpage
\section{Visualization of Generated Trajectories}\label{appendix:visual_mpe_all}
We provide additional visualization examples across various tasks (Spread and Tag). Each row corresponds to one of eight rollout samples. The leftmost column shows the original trajectory from the dataset, while the three middle columns illustrate intermediate steps of agent-wise sequential generation. The rightmost column (highlighted in red) presents the final trajectory obtained after all agents have been sequentially generated using subgoal-conditioned inpainting. These examples demonstrate that MASTARS produces cooperative, diverse, and globally coherent trajectories, contributing to improved performance through effective augmentation.

\begin{figure}[htbp]
    \centering
    \includegraphics[width=0.95\linewidth]{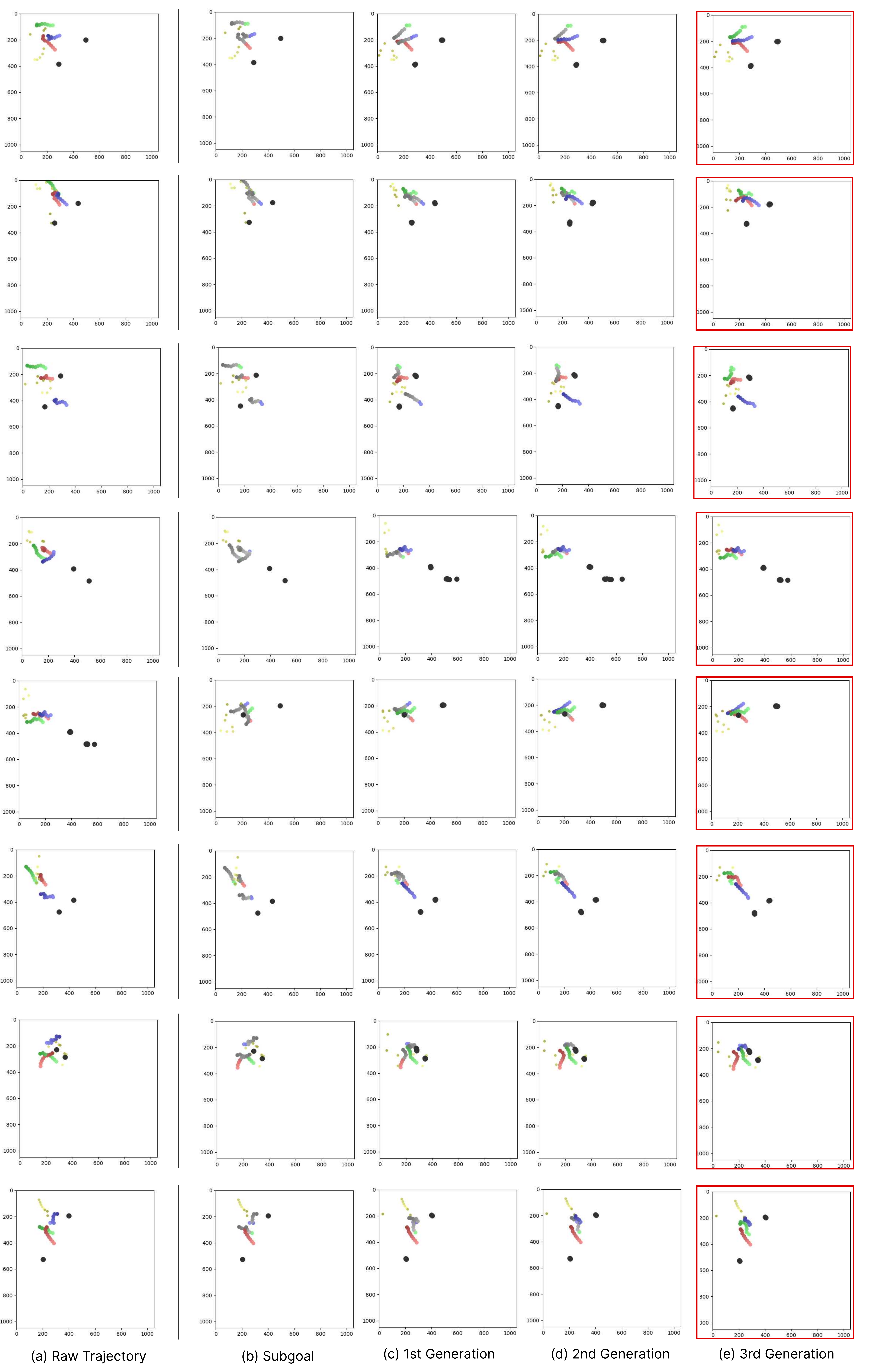}
    \caption{Visualization of the full trajectory generation process using our MASTARS framework in the \texttt{Tag-Medium} task. \texttt{X} marks are landmark positions reconstructed from each agent's observation.}
    \label{fig:appendix_tag_trajectory_viz}
\end{figure}

\begin{figure}[htbp]
    \centering
    \includegraphics[width=0.95\linewidth]{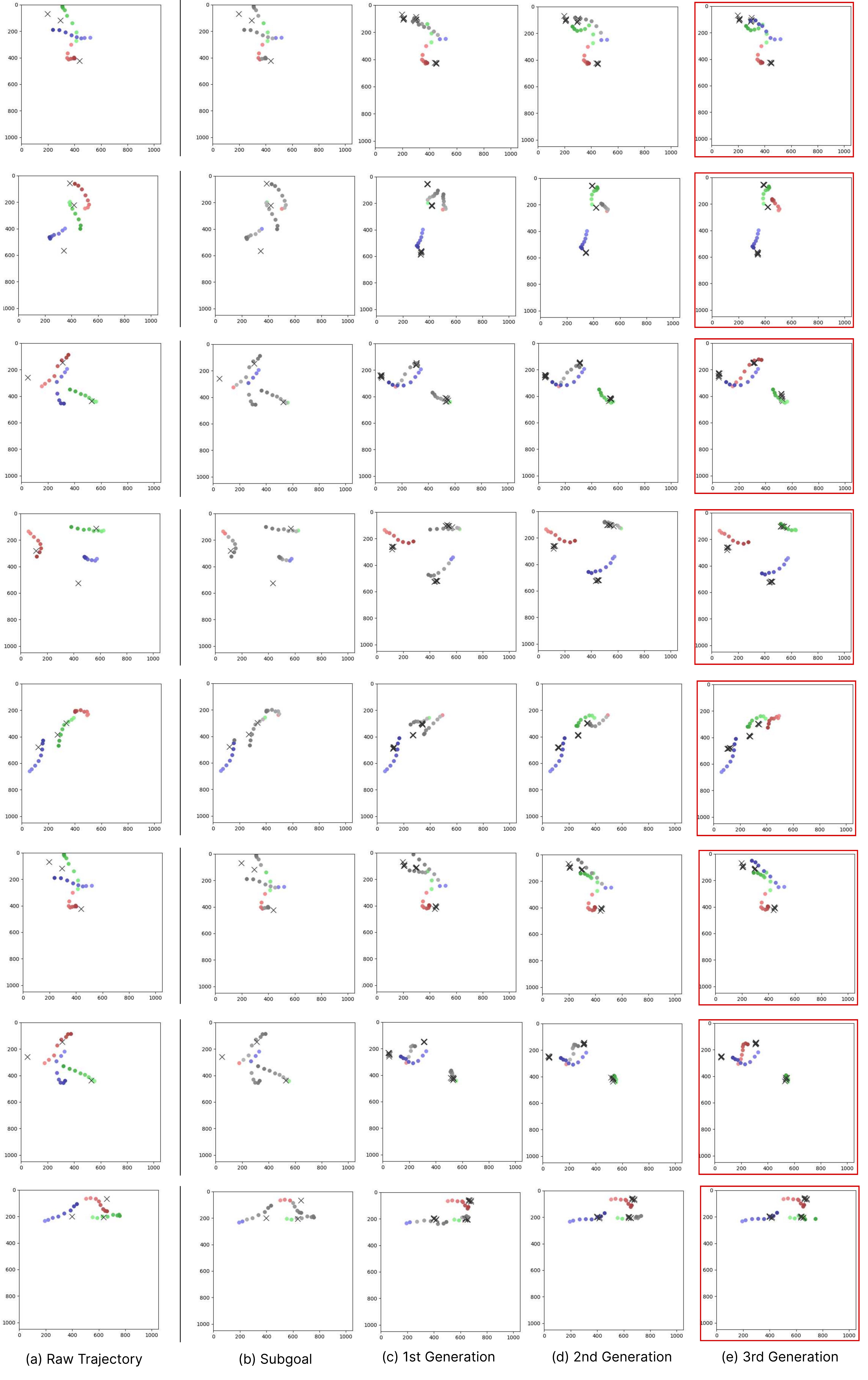}
    \caption{Visualization of the full trajectory generation process using our MASTARS framework in the \texttt{Spread-Medium} task. \texttt{X} marks are landmark positions reconstructed from each agent's observation.}
    \label{fig:appendix_spread_trajectory_viz}
\end{figure}

\newpage

\newpage
\section{Visual Comparison between MADiTS and Our Method}\label{appendix:visual_madits}
To further analyze the differences between our method and other joint generation approaches, we provide a visual comparison with MADiTS. Although the main paper includes comparisons against MADiff*, it is important to clarify that MADiff is not a trajectory generation method. To the best of our knowledge, among diffusion-based approaches applicable to multi-agent offline reinforcement learning, only two methods are designed to synthesize new data: MADiTS and INS. We therefore compare to MADiTS, which is the only prior work that performs genuine joint trajectory generation. We do not compare against INS, since INS produces transitions rather than trajectories.

\begin{figure}[ht]
	\centering
	\includegraphics[width=.65\linewidth]{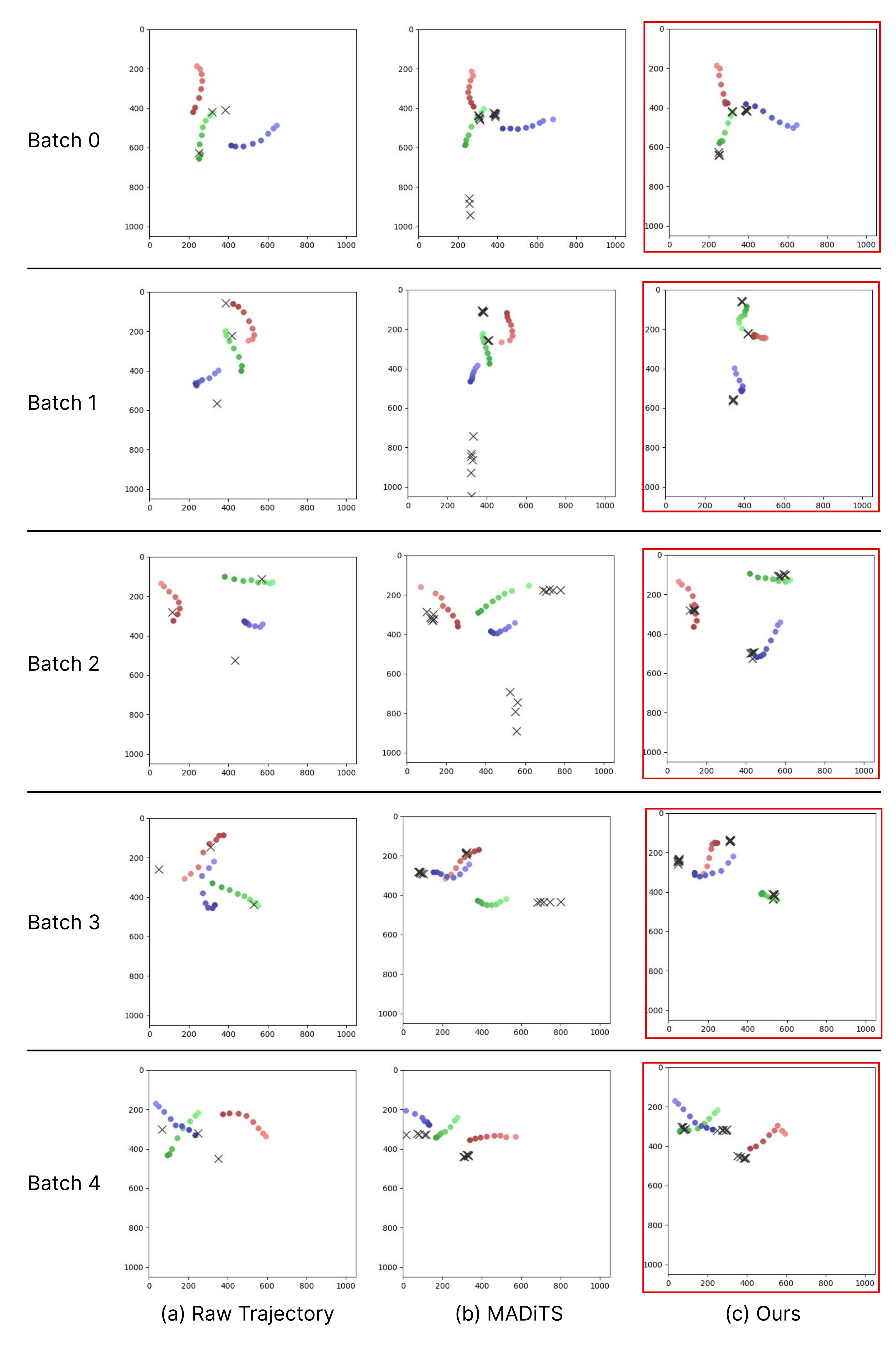}
	\caption{
        (a) Raw trajectory, (b) MADiTS, and (c) Ours}
	\label{fig:appendix_madits_and_us}
\end{figure}

As shown in Fig.~\ref{fig:appendix_madits_and_us}, MADiTS frequently fails to generate valid destination points (marked with X in the figure) and in some cases even fails to produce a feasible trajectory. This discrepancy stems from the fundamental differences in generation mechanisms. MADiTS relies on joint conditional diffusion to produce an entire trajectory in a single pass; as a result, its one-shot refinement process often lacks consistency, making the generated destinations noisy and poorly aligned with the surrounding agent behaviors.

In contrast, MASTARS adopts a RePaint-style inpainting strategy. By iteratively resampling masked segments conditioned on neighboring agents and previous denoising steps, our approach progressively corrects local inconsistencies and produces coherent, well-structured trajectories. This iterative refinement allows MASTARS to generate reliable and harmonized trajectories that MADiTS frequently fails to achieve.

\newpage


\newpage
\section*{NeurIPS Paper Checklist}

\begin{enumerate}

\item {\bf Claims}
    \item[] Question: Do the main claims made in the abstract and introduction accurately reflect the paper's contributions and scope?
    \item[] Answer: \answerYes{} 
    \item[] Justification:  The abstract and Section~\ref{introduction} clearly state the paper's main contributions, which are supported by the experimental results presented in Section~\ref{experiments}.
    \item[] Guidelines:
    \begin{itemize}
        \item The answer \answerNA{} means that the abstract and introduction do not include the claims made in the paper.
        \item The abstract and/or introduction should clearly state the claims made, including the contributions made in the paper and important assumptions and limitations. A \answerNo{} or \answerNA{} answer to this question will not be perceived well by the reviewers. 
        \item The claims made should match theoretical and experimental results, and reflect how much the results can be expected to generalize to other settings. 
        \item It is fine to include aspirational goals as motivation as long as it is clear that these goals are not attained by the paper. 
    \end{itemize}

\item {\bf Limitations}
    \item[] Question: Does the paper discuss the limitations of the work performed by the authors?
    \item[] Answer: \answerYes{} 
    \item[] Justification: We discuss the limitations and future directions of MASTARS in Appendix~\ref{appendix:limitation}, including the scalability of sequential generation with respect to the number of agents and potential extensions to subgoal generation.
    \item[] Guidelines:
    \begin{itemize}
        \item The answer \answerNA{} means that the paper has no limitation while the answer \answerNo{} means that the paper has limitations, but those are not discussed in the paper. 
        \item The authors are encouraged to create a separate ``Limitations'' section in their paper.
        \item The paper should point out any strong assumptions and how robust the results are to violations of these assumptions (e.g., independence assumptions, noiseless settings, model well-specification, asymptotic approximations only holding locally). The authors should reflect on how these assumptions might be violated in practice and what the implications would be.
        \item The authors should reflect on the scope of the claims made, e.g., if the approach was only tested on a few datasets or with a few runs. In general, empirical results often depend on implicit assumptions, which should be articulated.
        \item The authors should reflect on the factors that influence the performance of the approach. For example, a facial recognition algorithm may perform poorly when image resolution is low or images are taken in low lighting. Or a speech-to-text system might not be used reliably to provide closed captions for online lectures because it fails to handle technical jargon.
        \item The authors should discuss the computational efficiency of the proposed algorithms and how they scale with dataset size.
        \item If applicable, the authors should discuss possible limitations of their approach to address problems of privacy and fairness.
        \item While the authors might fear that complete honesty about limitations might be used by reviewers as grounds for rejection, a worse outcome might be that reviewers discover limitations that aren't acknowledged in the paper. The authors should use their best judgment and recognize that individual actions in favor of transparency play an important role in developing norms that preserve the integrity of the community. Reviewers will be specifically instructed to not penalize honesty concerning limitations.
    \end{itemize}

\item {\bf Theory assumptions and proofs}
    \item[] Question: For each theoretical result, does the paper provide the full set of assumptions and a complete (and correct) proof?
    \item[] Answer: \answerNA{} 
    \item[] Justification: The paper does not include theoretical results; all contributions are empirical.
    \item[] Guidelines:
    \begin{itemize}
        \item The answer \answerNA{} means that the paper does not include theoretical results. 
        \item All the theorems, formulas, and proofs in the paper should be numbered and cross-referenced.
        \item All assumptions should be clearly stated or referenced in the statement of any theorems.
        \item The proofs can either appear in the main paper or the supplemental material, but if they appear in the supplemental material, the authors are encouraged to provide a short proof sketch to provide intuition. 
        \item Inversely, any informal proof provided in the core of the paper should be complemented by formal proofs provided in appendix or supplemental material.
        \item Theorems and Lemmas that the proof relies upon should be properly referenced. 
    \end{itemize}

\item {\bf Experimental result reproducibility}
    \item[] Question: Does the paper fully disclose all the information needed to reproduce the main experimental results of the paper to the extent that it affects the main claims and/or conclusions of the paper (regardless of whether the code and data are provided or not)?
    \item[] Answer: \answerYes{} 
    \item[] Justification: The full algorithm of MASTARS is described in Section~\ref{method} and Appendix~\ref{appendix:pseudocode} (pseudocode), with training losses in Appendix~\ref{appendix:training_loss}. Detailed experimental settings, dataset compositions, and all hyperparameters (including post-processing thresholds) are provided in Appendices~\ref{appendix:experiment_detail} and \ref{appendix:implementation_detail}.
    \item[] Guidelines:
    \begin{itemize}
        \item The answer \answerNA{} means that the paper does not include experiments.
        \item If the paper includes experiments, a \answerNo{} answer to this question will not be perceived well by the reviewers: Making the paper reproducible is important, regardless of whether the code and data are provided or not.
        \item If the contribution is a dataset and\slash or model, the authors should describe the steps taken to make their results reproducible or verifiable. 
        \item Depending on the contribution, reproducibility can be accomplished in various ways. For example, if the contribution is a novel architecture, describing the architecture fully might suffice, or if the contribution is a specific model and empirical evaluation, it may be necessary to either make it possible for others to replicate the model with the same dataset, or provide access to the model. In general. releasing code and data is often one good way to accomplish this, but reproducibility can also be provided via detailed instructions for how to replicate the results, access to a hosted model (e.g., in the case of a large language model), releasing of a model checkpoint, or other means that are appropriate to the research performed.
        \item While NeurIPS does not require releasing code, the conference does require all submissions to provide some reasonable avenue for reproducibility, which may depend on the nature of the contribution. For example
        \begin{enumerate}
            \item If the contribution is primarily a new algorithm, the paper should make it clear how to reproduce that algorithm.
            \item If the contribution is primarily a new model architecture, the paper should describe the architecture clearly and fully.
            \item If the contribution is a new model (e.g., a large language model), then there should either be a way to access this model for reproducing the results or a way to reproduce the model (e.g., with an open-source dataset or instructions for how to construct the dataset).
            \item We recognize that reproducibility may be tricky in some cases, in which case authors are welcome to describe the particular way they provide for reproducibility. In the case of closed-source models, it may be that access to the model is limited in some way (e.g., to registered users), but it should be possible for other researchers to have some path to reproducing or verifying the results.
        \end{enumerate}
    \end{itemize}

\item {\bf Open access to data and code}
    \item[] Question: Does the paper provide open access to the data and code, with sufficient instructions to faithfully reproduce the main experimental results, as described in supplemental material?
    \item[] Answer: \answerYes{} 
    \item[] Justification: The source code, along with instructions to reproduce the main experimental results, is publicly available at \url{https://github.com/Jiwonjeon9603/MASTARS}. The benchmark datasets used in this paper (MPE, SMAC, og-marl) are publicly available, as cited in Appendix~\ref{appendix:experiment_detail}.
    \item[] Guidelines:
    \begin{itemize}
        \item The answer \answerNA{} means that paper does not include experiments requiring code.
        \item Please see the NeurIPS code and data submission guidelines (\url{https://neurips.cc/public/guides/CodeSubmissionPolicy}) for more details.
        \item While we encourage the release of code and data, we understand that this might not be possible, so \answerNo{} is an acceptable answer. Papers cannot be rejected simply for not including code, unless this is central to the contribution (e.g., for a new open-source benchmark).
        \item The instructions should contain the exact command and environment needed to run to reproduce the results. See the NeurIPS code and data submission guidelines (\url{https://neurips.cc/public/guides/CodeSubmissionPolicy}) for more details.
        \item The authors should provide instructions on data access and preparation, including how to access the raw data, preprocessed data, intermediate data, and generated data, etc.
        \item The authors should provide scripts to reproduce all experimental results for the new proposed method and baselines. If only a subset of experiments are reproducible, they should state which ones are omitted from the script and why.
        \item At submission time, to preserve anonymity, the authors should release anonymized versions (if applicable).
        \item Providing as much information as possible in supplemental material (appended to the paper) is recommended, but including URLs to data and code is permitted.
    \end{itemize}

\item {\bf Experimental setting/details}
    \item[] Question: Does the paper specify all the training and test details (e.g., data splits, hyperparameters, how they were chosen, type of optimizer) necessary to understand the results?
    \item[] Answer: \answerYes{} 
    \item[] Justification: All training and evaluation details, including dataset compositions, algorithm configurations, network architectures, training schedules, and task-specific hyperparameters such as post-processing thresholds, are provided in Appendices~\ref{appendix:experiment_detail} and \ref{appendix:implementation_detail}.
    \item[] Guidelines:
    \begin{itemize}
        \item The answer \answerNA{} means that the paper does not include experiments.
        \item The experimental setting should be presented in the core of the paper to a level of detail that is necessary to appreciate the results and make sense of them.
        \item The full details can be provided either with the code, in appendix, or as supplemental material.
    \end{itemize}

\item {\bf Experiment statistical significance}
    \item[] Question: Does the paper report error bars suitably and correctly defined or other appropriate information about the statistical significance of the experiments?
    \item[] Answer: \answerYes{} 
    \item[] Justification: All main results are reported as mean $\pm$ standard deviation across five random seeds, as described in Appendix~\ref{appendix:implementation_detail}.
    \item[] Guidelines:
    \begin{itemize}
        \item The answer \answerNA{} means that the paper does not include experiments.
        \item The authors should answer \answerYes{} if the results are accompanied by error bars, confidence intervals, or statistical significance tests, at least for the experiments that support the main claims of the paper.
        \item The factors of variability that the error bars are capturing should be clearly stated (for example, train/test split, initialization, random drawing of some parameter, or overall run with given experimental conditions).
        \item The method for calculating the error bars should be explained (closed form formula, call to a library function, bootstrap, etc.)
        \item The assumptions made should be given (e.g., Normally distributed errors).
        \item It should be clear whether the error bar is the standard deviation or the standard error of the mean.
        \item It is OK to report 1-sigma error bars, but one should state it. The authors should preferably report a 2-sigma error bar than state that they have a 96\% CI, if the hypothesis of Normality of errors is not verified.
        \item For asymmetric distributions, the authors should be careful not to show in tables or figures symmetric error bars that would yield results that are out of range (e.g., negative error rates).
        \item If error bars are reported in tables or plots, the authors should explain in the text how they were calculated and reference the corresponding figures or tables in the text.
    \end{itemize}

\item {\bf Experiments compute resources}
    \item[] Question: For each experiment, does the paper provide sufficient information on the computer resources (type of compute workers, memory, time of execution) needed to reproduce the experiments?
    \item[] Answer: \answerYes{} 
    \item[] Justification: The hardware specifications (8 NVIDIA GeForce RTX 3090 GPUs, 96 logical CPU cores) and runtime comparisons between MASTARS and baselines are reported in Appendix~\ref{appendix:computational_cost}.
    \item[] Guidelines:
    \begin{itemize}
        \item The answer \answerNA{} means that the paper does not include experiments.
        \item The paper should indicate the type of compute workers CPU or GPU, internal cluster, or cloud provider, including relevant memory and storage.
        \item The paper should provide the amount of compute required for each of the individual experimental runs as well as estimate the total compute. 
        \item The paper should disclose whether the full research project required more compute than the experiments reported in the paper (e.g., preliminary or failed experiments that didn't make it into the paper). 
    \end{itemize}
    
\item {\bf Code of ethics}
    \item[] Question: Does the research conducted in the paper conform, in every respect, with the NeurIPS Code of Ethics \url{https://neurips.cc/public/EthicsGuidelines}?
    \item[] Answer: \answerYes{} 
    \item[] Justification: We have reviewed the NeurIPS Code of Ethics, and our research conforms to it in every respect.
    \item[] Guidelines:
    \begin{itemize}
        \item The answer \answerNA{} means that the authors have not reviewed the NeurIPS Code of Ethics.
        \item If the authors answer \answerNo, they should explain the special circumstances that require a deviation from the Code of Ethics.
        \item The authors should make sure to preserve anonymity (e.g., if there is a special consideration due to laws or regulations in their jurisdiction).
    \end{itemize}

\item {\bf Broader impacts}
    \item[] Question: Does the paper discuss both potential positive societal impacts and negative societal impacts of the work performed?
    \item[] Answer: \answerYes{} 
    \item[] Justification: We discuss the broader impacts of MASTARS in Appendix~\ref{appendix:broader_impact}, including its potential to improve sample efficiency in safety-critical applications such as autonomous driving and robotics, as well as the need for careful validation when deploying in such environments.
    \item[] Guidelines:
    \begin{itemize}
        \item The answer \answerNA{} means that there is no societal impact of the work performed.
        \item If the authors answer \answerNA{} or \answerNo, they should explain why their work has no societal impact or why the paper does not address societal impact.
        \item Examples of negative societal impacts include potential malicious or unintended uses (e.g., disinformation, generating fake profiles, surveillance), fairness considerations (e.g., deployment of technologies that could make decisions that unfairly impact specific groups), privacy considerations, and security considerations.
        \item The conference expects that many papers will be foundational research and not tied to particular applications, let alone deployments. However, if there is a direct path to any negative applications, the authors should point it out. For example, it is legitimate to point out that an improvement in the quality of generative models could be used to generate Deepfakes for disinformation. On the other hand, it is not needed to point out that a generic algorithm for optimizing neural networks could enable people to train models that generate Deepfakes faster.
        \item The authors should consider possible harms that could arise when the technology is being used as intended and functioning correctly, harms that could arise when the technology is being used as intended but gives incorrect results, and harms following from (intentional or unintentional) misuse of the technology.
        \item If there are negative societal impacts, the authors could also discuss possible mitigation strategies (e.g., gated release of models, providing defenses in addition to attacks, mechanisms for monitoring misuse, mechanisms to monitor how a system learns from feedback over time, improving the efficiency and accessibility of ML).
    \end{itemize}
    
\item {\bf Safeguards}
    \item[] Question: Does the paper describe safeguards that have been put in place for responsible release of data or models that have a high risk for misuse (e.g., pre-trained language models, image generators, or scraped datasets)?
    \item[] Answer: \answerNA{} 
    \item[] Justification: This work does not release any pre-trained models, datasets, or other artifacts with a high risk of misuse. The augmented data is generated for standard simulated benchmarks (MPE, SMAC, SMAC-v2) and poses no such risks.
    \item[] Guidelines:
    \begin{itemize}
        \item The answer \answerNA{} means that the paper poses no such risks.
        \item Released models that have a high risk for misuse or dual-use should be released with necessary safeguards to allow for controlled use of the model, for example by requiring that users adhere to usage guidelines or restrictions to access the model or implementing safety filters. 
        \item Datasets that have been scraped from the Internet could pose safety risks. The authors should describe how they avoided releasing unsafe images.
        \item We recognize that providing effective safeguards is challenging, and many papers do not require this, but we encourage authors to take this into account and make a best faith effort.
    \end{itemize}

\item {\bf Licenses for existing assets}
    \item[] Question: Are the creators or original owners of assets (e.g., code, data, models), used in the paper, properly credited and are the license and terms of use explicitly mentioned and properly respected?
    \item[] Answer: \answerYes{} 
    \item[] Justification: All existing assets used in this paper, including the MPE, SMAC, SMAC-v2 benchmarks and the og-marl, PyMARL, and QMIX implementations, are properly cited and used in accordance with their respective licenses.
    \item[] Guidelines:
    \begin{itemize}
        \item The answer \answerNA{} means that the paper does not use existing assets.
        \item The authors should cite the original paper that produced the code package or dataset.
        \item The authors should state which version of the asset is used and, if possible, include a URL.
        \item The name of the license (e.g., CC-BY 4.0) should be included for each asset.
        \item For scraped data from a particular source (e.g., website), the copyright and terms of service of that source should be provided.
        \item If assets are released, the license, copyright information, and terms of use in the package should be provided. For popular datasets, \url{paperswithcode.com/datasets} has curated licenses for some datasets. Their licensing guide can help determine the license of a dataset.
        \item For existing datasets that are re-packaged, both the original license and the license of the derived asset (if it has changed) should be provided.
        \item If this information is not available online, the authors are encouraged to reach out to the asset's creators.
    \end{itemize}

\item {\bf New assets}
    \item[] Question: Are new assets introduced in the paper well documented and is the documentation provided alongside the assets?
    \item[] Answer: \answerYes{} 
    \item[] Justification: The MASTARS code is publicly released with documentation at \url{https://github.com/Jiwonjeon9603/MASTARS}. The data generation procedure for our custom SMAC-v2 datasets, including unit-type sampling distributions and dataset statistics, is fully described in Appendix~\ref{appendix:smacv2}.
    \item[] Guidelines:
    \begin{itemize}
        \item The answer \answerNA{} means that the paper does not release new assets.
        \item Researchers should communicate the details of the dataset\slash code\slash model as part of their submissions via structured templates. This includes details about training, license, limitations, etc. 
        \item The paper should discuss whether and how consent was obtained from people whose asset is used.
        \item At submission time, remember to anonymize your assets (if applicable). You can either create an anonymized URL or include an anonymized zip file.
    \end{itemize}

\item {\bf Crowdsourcing and research with human subjects}
    \item[] Question: For crowdsourcing experiments and research with human subjects, does the paper include the full text of instructions given to participants and screenshots, if applicable, as well as details about compensation (if any)? 
    \item[] Answer: \answerNA{} 
    \item[] Justification: This work does not involve crowdsourcing or research with human subjects.
    \item[] Guidelines:
    \begin{itemize}
        \item The answer \answerNA{} means that the paper does not involve crowdsourcing nor research with human subjects.
        \item Including this information in the supplemental material is fine, but if the main contribution of the paper involves human subjects, then as much detail as possible should be included in the main paper. 
        \item According to the NeurIPS Code of Ethics, workers involved in data collection, curation, or other labor should be paid at least the minimum wage in the country of the data collector. 
    \end{itemize}

\item {\bf Institutional review board (IRB) approvals or equivalent for research with human subjects}
    \item[] Question: Does the paper describe potential risks incurred by study participants, whether such risks were disclosed to the subjects, and whether Institutional Review Board (IRB) approvals (or an equivalent approval/review based on the requirements of your country or institution) were obtained?
    \item[] Answer: \answerNA{} 
    \item[] Justification: This work does not involve research with human subjects, and IRB approval is therefore not applicable.
    \item[] Guidelines:
    \begin{itemize}
        \item The answer \answerNA{} means that the paper does not involve crowdsourcing nor research with human subjects.
        \item Depending on the country in which research is conducted, IRB approval (or equivalent) may be required for any human subjects research. If you obtained IRB approval, you should clearly state this in the paper. 
        \item We recognize that the procedures for this may vary significantly between institutions and locations, and we expect authors to adhere to the NeurIPS Code of Ethics and the guidelines for their institution. 
        \item For initial submissions, do not include any information that would break anonymity (if applicable), such as the institution conducting the review.
    \end{itemize}

\item {\bf Declaration of LLM usage}
    \item[] Question: Does the paper describe the usage of LLMs if it is an important, original, or non-standard component of the core methods in this research? Note that if the LLM is used only for writing, editing, or formatting purposes and does \emph{not} impact the core methodology, scientific rigor, or originality of the research, declaration is not required.
    \item[] Answer: \answerNA{} 
    \item[] Justification: LLMs are not used as any component of the core methodology of this research, and were only used as assistive tools for writing and grammar checking, which does not require declaration per the NeurIPS LLM policy.
    \item[] Guidelines:
    \begin{itemize}
        \item The answer \answerNA{} means that the core method development in this research does not involve LLMs as any important, original, or non-standard components.
        \item Please refer to our LLM policy in the NeurIPS handbook for what should or should not be described.
    \end{itemize}
\end{enumerate}

\end{document}